\documentclass[10pt,twocolumn,letterpaper]{article}

\usepackage[pagenumbers]{cvpr} 

\usepackage{graphicx}
\usepackage{amsmath}
\usepackage{amssymb}
\usepackage{booktabs}

\usepackage[utf8]{inputenc} 
\usepackage[T1]{fontenc}    
\usepackage{url}            
\usepackage{amsfonts}       
\usepackage{nicefrac}       
\usepackage{microtype}      
\usepackage{xcolor}         
\usepackage{multirow}
\usepackage{comment}
\usepackage{soul}
\usepackage{caption}[font=small]
\usepackage{subcaption}
\definecolor{applegreen}{rgb}{0.55, 0.71, 0.0}
\definecolor{alizarin}{rgb}{0.82, 0.1, 0.26}
\usepackage[normalem]{ulem}
\newif\ifmodify
\modifytrue 
\ifmodify
\newcommand{\dl}[1]{\textcolor{red}{\sout{#1}}}
\newcommand{\f}[1]{\textcolor{blue}{#1}}

\else
\newcommand{\dl}[1]{}
\newcommand{\f}[1]{#1}

\fi

\usepackage[pagebackref,breaklinks,colorlinks]{hyperref}

\usepackage[capitalize]{cleveref}
\crefname{section}{Sec.}{Secs.}
\Crefname{section}{Section}{Sections}
\Crefname{table}{Table}{Tables}
\crefname{table}{Tab.}{Tabs.}

\def\cvprPaperID{12513} 
\def\confName{CVPR}
\def\confYear{2023}

\begin{document}

\title{MILAN: \underline{M}asked \underline{I}mage Pretraining on \underline{L}anguage \underline{A}ssisted Representatio\underline{n}}


\author{
Zejiang Hou$^1$
\and
Fei Sun$^2$
\and
Yen-Kuang Chen$^2$
\and
Yuan Xie$^2$
\and
Sun-Yuan Kung$^1$
\and
\centerline{$^1$Princeton University~~$^2$DAMO Academy, Alibaba Group}
}

\maketitle

\begin{abstract}
Self-attention based transformer models have been dominating many computer vision tasks in the past few years. 
Their superb model qualities 
{heavily depend on the} excessively large labeled image datasets.
In order to reduce the reliance on large labeled datasets, reconstruction based 
masked autoencoders are gaining popularity, which learn high quality transferable representations from unlabeled images.
For the same purpose, recent weakly supervised image pretraining methods explore language supervision from text captions accompanying the images.
In this work, we propose masked image pretraining on language assisted representation, dubbed as MILAN.
Instead of predicting raw pixels or low level features, our pretraining objective is to reconstruct the image features with substantial semantic signals that are obtained using caption supervision.
Moreover, to accommodate our reconstruction target, we propose a more efficient {prompting} decoder architecture and
{a semantic aware mask sampling mechanism}, 
which further advance the transfer performance of the pretrained model. 
Experimental results demonstrate that MILAN
delivers higher accuracy than the previous works.
When the masked autoencoder is pretrained and finetuned on ImageNet-1K dataset with an input resolution of 224$\times$224, MILAN achieves a top-1 accuracy of 85.4\% {on} ViT-Base, surpassing
previous state-of-the-arts by 1\%.
In the downstream
semantic segmentation task, MILAN achieves
52.7 mIoU using ViT-Base on ADE20K dataset, outperforming previous masked pretraining results by 4 points\footnote{Code is available at \url{https://github.com/zejiangh/MILAN}.}.
\end{abstract}

\section{Introduction}

In recent years, we have seen a wide adoption of applying natural language processing (NLP) techniques in computer vision (CV) tasks. The vision transformer (ViT) model~\cite{dosovitskiy2020vit} applies the self-attention based transformer architecture to vision tasks and {have} achieved remarkable performance.
However, training 
{ViT models} 
requires much larger labeled datasets to avoid overfitting,
such as ImageNet-22K~\cite{deng2009imagenet} and JFT-300M~\cite{sun2017jft}.
Explicitly labeling large image datasets is hardly affordable.

Reconstruction based self-supervised pretraining can extract semantic information from unlabeled data, and has become a popular method to reduce the reliance on very large labeled datasets in both NLP and CV.
It is first exemplified by BERT~\cite{devlin2018bert} in NLP.
Acting like a masked autoencoder~\cite{vincent2010denoising}, BERT randomly masks some percentage of the input word tokens and learns to reconstruct the vocabularies of those masked tokens.
Works in~\cite{bao2021beit,he2021mae,xie2021simmim,wei2021maskfeat} adopt similar techniques in CV to address the data-hungry issue of ViT models. A {large} percentage of the input image patches are randomly masked with the goal of reconstructing them. 

The mask autoencoders can be extended in several directions. 
Unlike the masked word tokens in NLP, which contain rich semantic information, the masked image patches only contain low-level pixel data.
Several works~\cite{bao2021beit, dong2021peco, chen2022cae} explore more abstract \textbf{reconstruction targets}, aiming to learn higher level visual concepts. However, those methods still only retrieve semantic signals from raw image pixels, which by itself is a difficult task.
In addition, the selection of the reconstruction targets heavily influences the \textbf{decoder design} in autoencoders, as the decoder serves to reconstruct the masked features with the guidance from the encoder's output representations.
Full fledged transformer blocks are used in {the} decoder of MAE~\cite{he2021mae} to reconstruct masked input patches pixel by pixel, whereas lightweight linear layer is adopted in {the decoder of} MaskFeat~\cite{wei2021maskfeat} to reconstruct local features of the image. 
Thus, if we were using more semantic preserving reconstruction targets, a task tailored decoder architecture would be required.
Furthermore, different \textbf{sampling strategies} (\textit{e.g.,} grid, block, random) of the input image patches affect the final performance of masked image pretraining~\cite{he2021mae, xie2021simmim}.
Majority of prior arts~\cite{bao2021beit, dong2021peco, he2021mae, xie2021simmim, wei2021maskfeat, baevski2022data2vec, chen2022cae} sample the masked patches uniformly at random
since it is unbiased and can guarantee coverage. 
However, it is indifferent to more discriminative image patches and unimportant ones, thus may suffer from slow training convergence \cite{kakogeorgiou2022attentionmask}.

\begin{figure*}[t]
\centering
\includegraphics[scale=0.16]{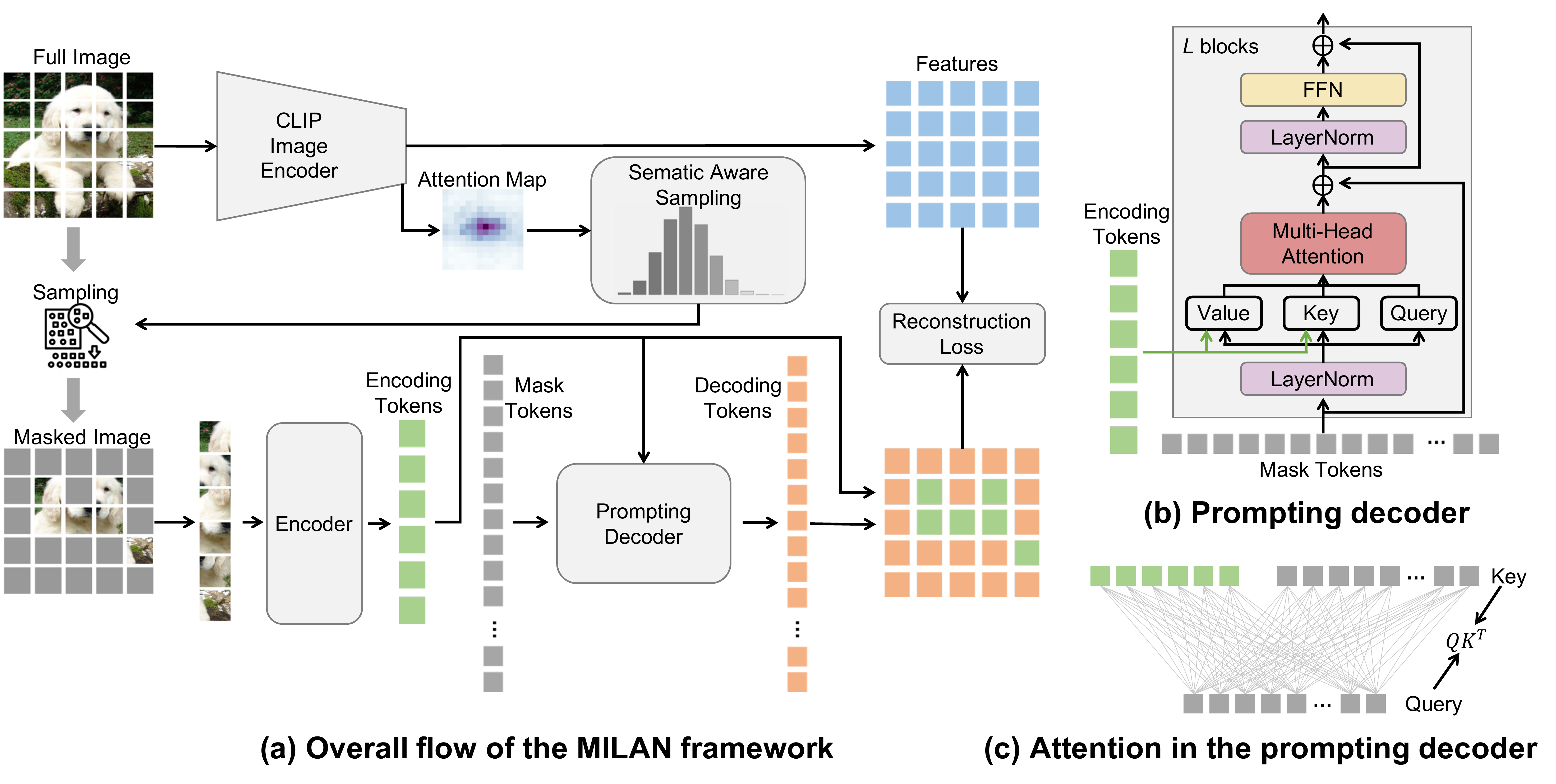}
\vspace{-0.1in}
\caption{(a) The overall flow of MILAN. The masked autoencoder uses the outputs of the CLIP model as the reconstruction target. An efficient prompting decoder freezes the features of the encoding tokens and only updates the mask tokens. 
{A semantic aware sampling} is used to guide the selection of the unmasked image patches. {The reconstruction loss is computed on the representation features of both masked and unmasked patches}.
(b) A detailed diagram of the {prompting} decoder. (c) The attention computation in the {prompting} decoder.}
\label{fig:architecture_overview}
\end{figure*}

In this work, we analyze three highly correlated aspects in masked autoencoders: the reconstruction target, the decoder design, and the mask sampling strategy.
We propose a new approach called MILAN, which performs masked image pretraining on language assisted representations. In specific: 

(1) We recognize the limitation of extracting semantic signals from raw image pixels alone. But such signals are readily available in the captions accompanying the images.
Recent works such as CLIP~\cite{radford2021clip}, SLIP~\cite{mu2021slip}, and COCA~\cite{yu2022coca} explore the use of caption supervision to learn image representations on abundant image-text pairs obtained from the Internet.
The output image features from those models implicitly contain semantic information that facilitates the interpretation of the image contents.
In this work, we take the image features coming out of the CLIP image encoder \cite{radford2021clip} as the reconstruction targets for the masked image pretraining, which benefits from natural language supervision and encourages the model to learn high level visual concepts.
More interestingly, we will show that the quality of the representation improves on the targets after masked image pretraining.

(2) We realize the tight coupling between the decoder architecture and the reconstruction targets.
We design an efficient prompting decoder suitable for reconstruction targets that are latent representations containing affluent semantic signals.
It freezes the encoder's output representations of the unmasked patches and uses them as ``fixed prompts'' to reconstruct the features of the masked patches. 
Prompting decoder achieves higher accuracy and reduces the decoding computational cost simultaneously.

(3) Different image patch sampling strategies impact pretraining efficiency.
Since our reconstruction targets provide global structure information of the images,
we propose a semantic aware mask sampling mechanism to discriminate
semantically important image patches 
from the insignificant background patches, 
which improves representation quality and pretraining efficiency. 

(4) Combining the three aspects leads to our MILAN framework (Figure \ref{fig:architecture_overview}). 
Experimentally, our ViT-Base and ViT-Large model{s} pretrained and finetuned on ImageNet-1K dataset achieve 86.4\% and 88.3\% top-1 accuracy, respectively.
Moreover, MILAN significantly boosts the linear probing accuracy compared to
reconstruction based and language-image based pretraining methods, and achieves state-of-the-art performance on the downstream object detection, instance segmentation, and semantic segmentation tasks.

\begin{figure*}[t]
\begin{small}
     \centering
     \begin{subfigure}[b]{0.244\textwidth}
         \centering
         \includegraphics[width=\textwidth]{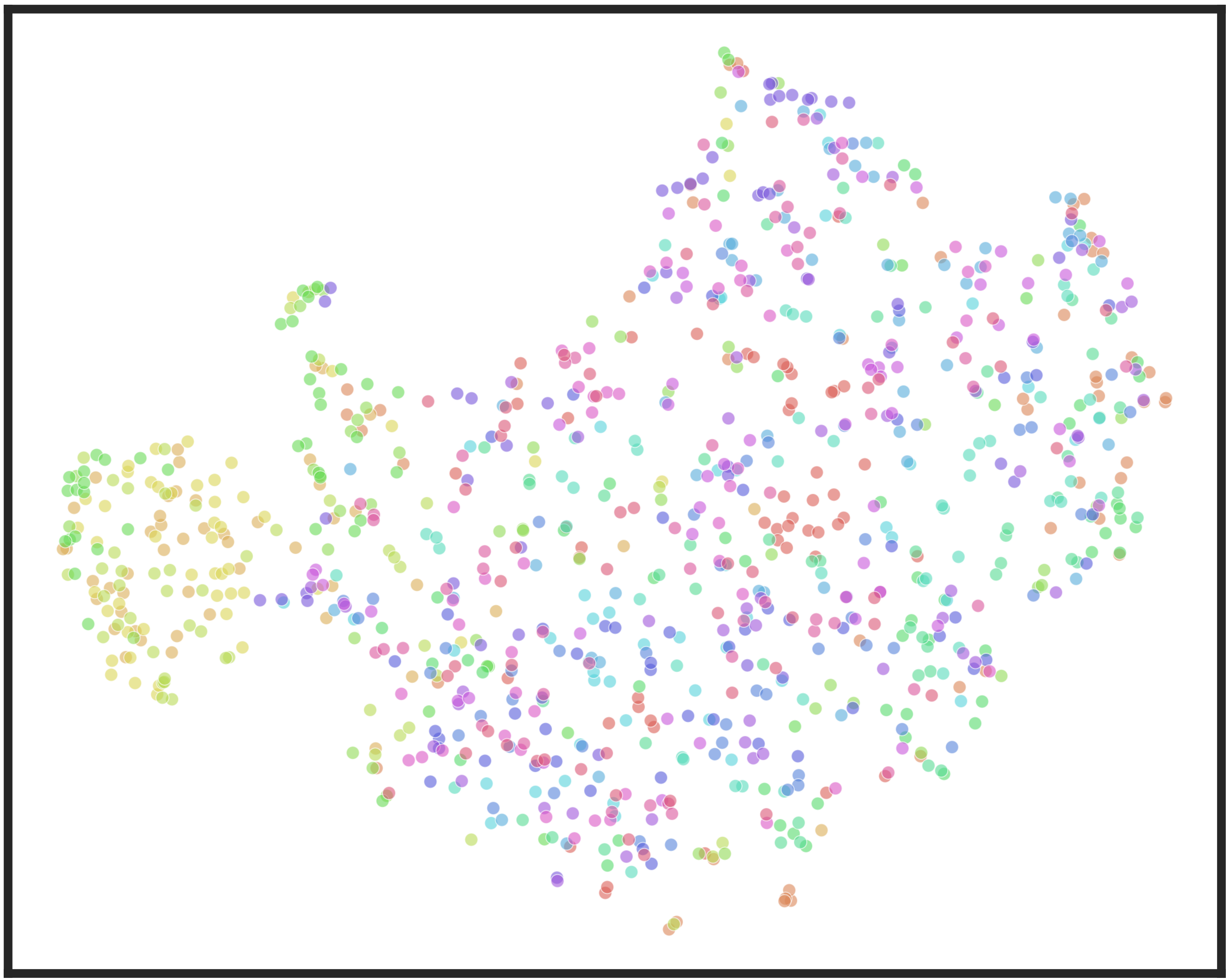}
         \caption{MAE pretrained}
     \end{subfigure}
     \hfill
     \begin{subfigure}[b]{0.244\textwidth}
         \centering
         \includegraphics[width=\textwidth]{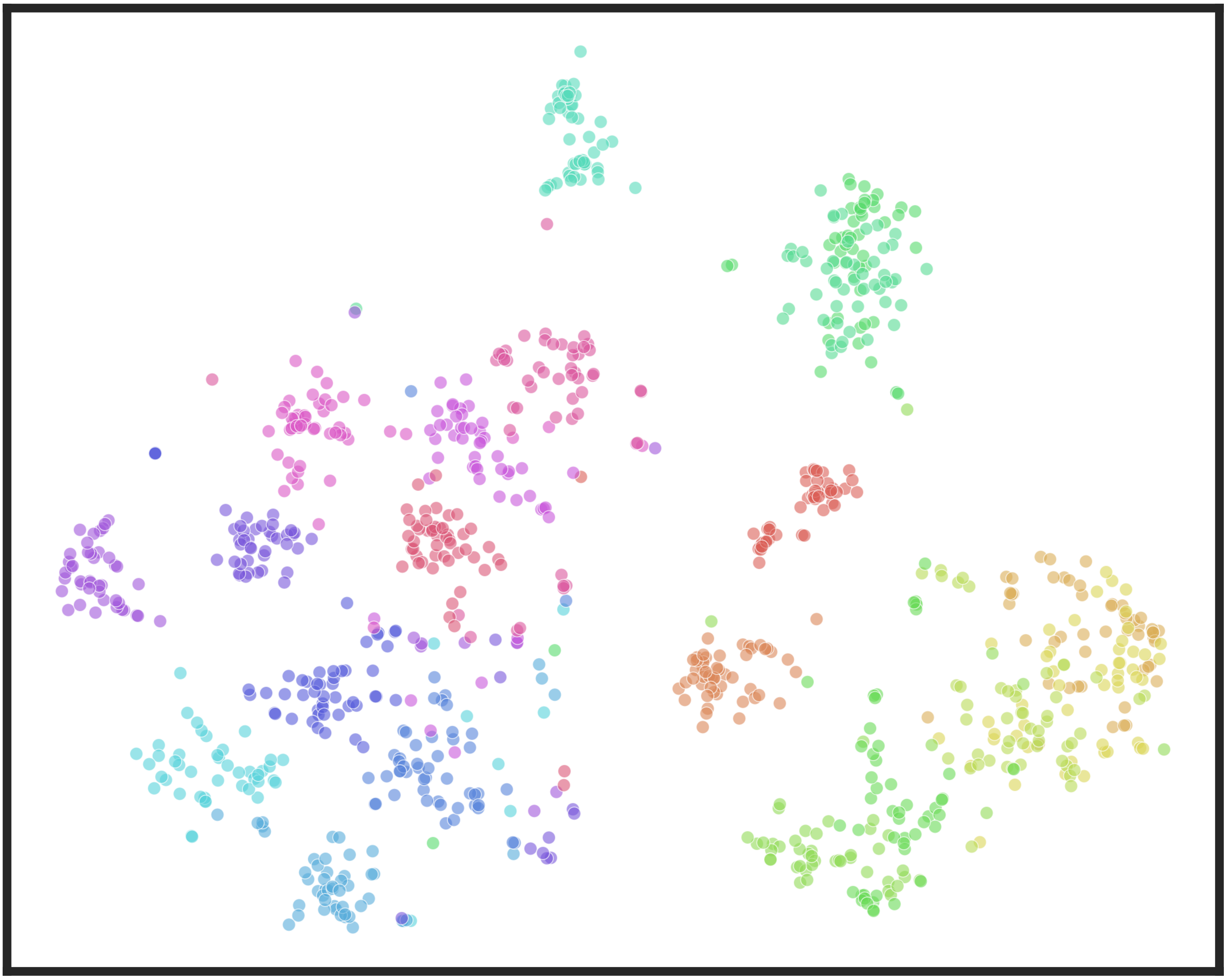}
         \caption{CLIP image encoder}
     \end{subfigure}
     \hfill
     \begin{subfigure}[b]{0.244\textwidth}
         \centering
         \includegraphics[width=\textwidth]{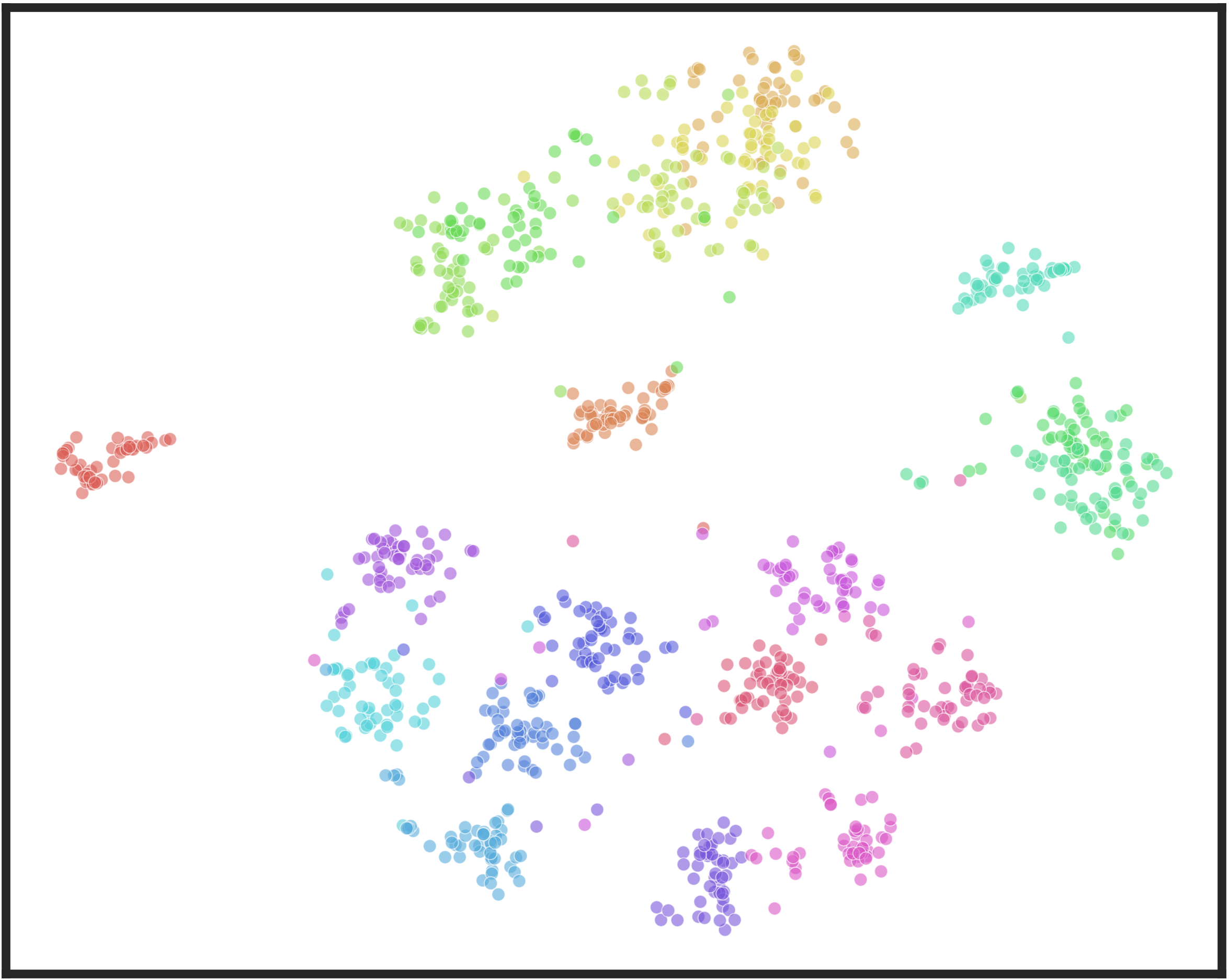}
         \caption{MILAN pretrained}
     \end{subfigure}
     \hfill
     \begin{subfigure}[b]{0.244\textwidth}
         \centering
         \includegraphics[width=\textwidth]{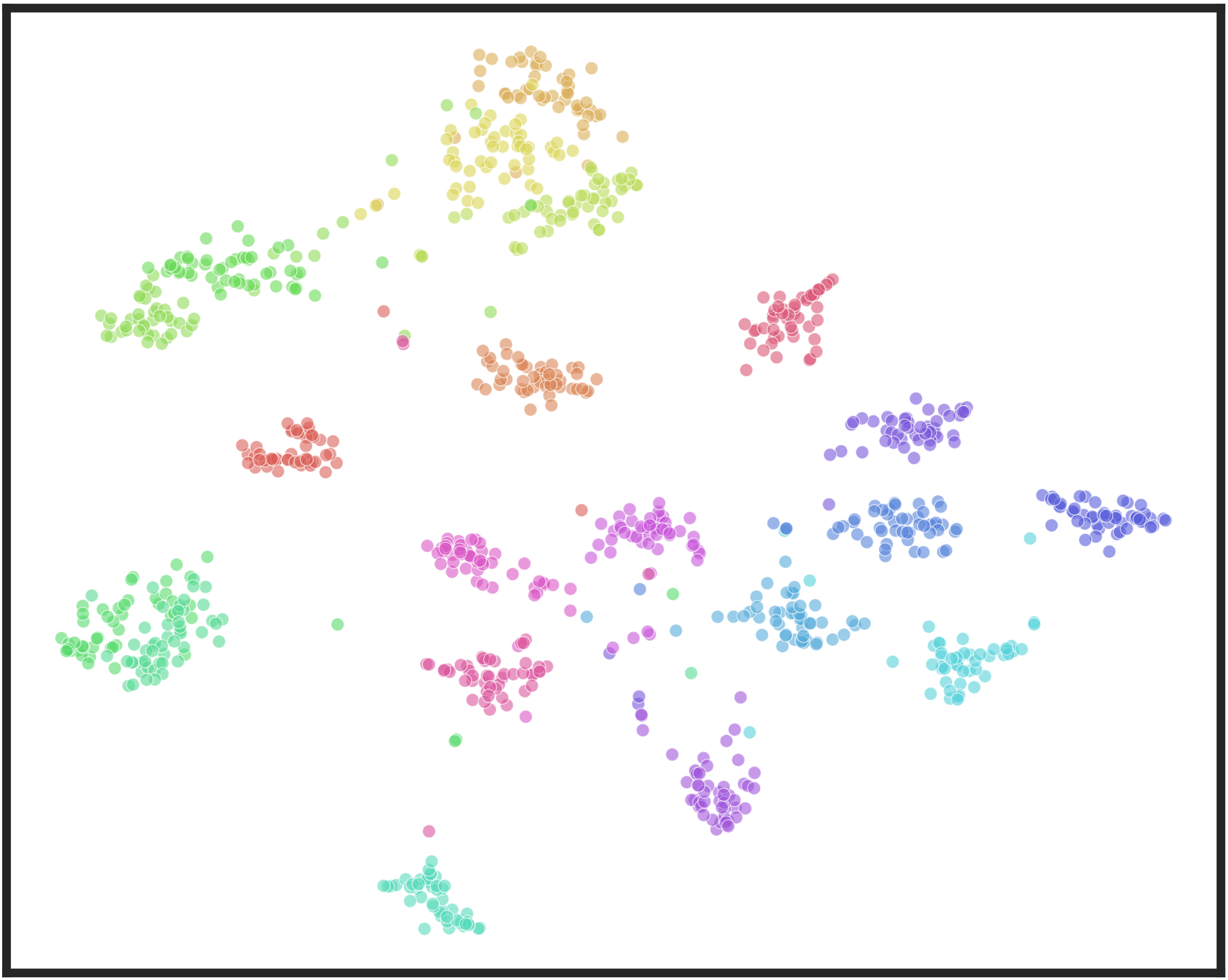}
         \caption{MILAN finetuned}
     \end{subfigure}
     \vspace{-0.1in}
        \caption{t-SNE visualization of the learned features from ViT-Base obtained by different pretraining methods. We plot the features before the final linear head.
        We use images of randomly sampled 20 classes in ImageNet-1K validation split.}
        \label{fig:tsne}
\end{small}     
\end{figure*}

\section{Methodology}

\subsection{Overview}

The overall flow of MILAN is illustrated in Figure \ref{fig:architecture_overview}(a).
We use a masked autoencoder architecture similar to MAE \cite{he2021mae}.
The encoder {transforms} the unmasked patches into latent representations.
The decoder reconstructs the representations of the masked patches assisted by the features of the unmasked patches. 
We use the latent features that the CLIP image encoder {produces} from the full image as the reconstruction targets, which are contextualized representations based on the global structure of the image and caption information.
The attention map is extracted from the last self-attention layer of the CLIP image encoder and is used to construct a semantic aware sampling distribution to sample unmasked patches.
The sampled patches are sent into the encoder and mapped to the latent feature space.
We design a prompting decoder that freezes the encoder's output when hallucinating the features of the masked patches from mask tokens.
As shown in Figure \ref{fig:architecture_overview}(b), the query of the attention block in the prompting decoder only contains the features of mask tokens.
The key and value matrices comprise both the encoder's outputs and the features of mask tokens.
The masked patches' features from the prompting decoder and the unmasked patches' features from the encoder are combined at the end. We re-order the combined full set of features to align with the targets, and compute the reconstruction loss.

 MILAN differs from its closely related MAE in: 1) the targets we predict are latent representations obtained with language guidance, whereas MAE reconstructs raw pixels; 2) mask sampling in MILAN is more adapted to patches' discriminativeness in contrast to MAE's uniform sampling; 3) our prompting decoder does not update the encoder's output and thus is more efficient.

\subsection{Reconstruction target: language assisted representation}
The reconstruction target is a crucial component in masked image pretraining. It influences the semantics of the learned latent representations. Language naturally contains rich semantics, while such information is more difficult to extract from {image} pixels directly. Thus, an image-text pair provides more meaningful learning {signals} than an image alone. In practice, texts accompanying images can be easily obtained at scale in the form of image captions. Such large unlabeled image-text datasets foster a series of weakly-supervised image pretraining methods \cite{radford2021clip,mu2021slip,li2021declip,singh2022swag} with caption supervision. It is expected that the visual representations learned with language guidance can provide affluent semantic information. Therefore, we take the language assisted representations as the reconstruction target in our masked image pretraining framework.

In this work, we primarily use the pretrained CLIP \cite{radford2021clip} model to generate the reconstruction targets. CLIP is trained on a dataset containing images and free-form text
captions with InfoNCE loss \cite{van2018cpc}. The model has an image encoder and a text encoder, both using the transformer architecture. The encoded image and text features are projected to the same dimension and normalized. Given a batch of image-text pairs, CLIP trains the image
encoder and the text encoder by maximizing the feature cosine similarity {for} the
{matching} image-text pairs in the batch while minimizing the cosine similarity {for} all {other} non-matching pairs.
Though without labeling data, the image features are trained to be close to the paired text features and distill the rich semantics embedded in the text features.
This is illustrated in Figure~\ref{fig:tsne}, where the last layer features produced by MAE's pretrained model, CLIP's image encoder, and our pretrained model are visualized by t-SNE plots.
As shown, the representations from CLIP's image encoder for each category tend to be grouped together while the pretrained model from MAE cannot distinguish the visual concepts in diffferent categories. MILAN adopts the image features from CLIP as the target and trains the model with a more challenging masked prediction objective. The learned representations are better clustered for different categories.

The pretraining objective of MILAN is formally described as follows.
Let $f_{\theta}$ denote the CLIP image encoder, whose weights $\theta$ are frozen. The masked autoencoder under training comprises encoder $g_{\xi}$ with weights $\xi$ and decoder $h_{\nu}$ with weights $\nu.$
From a given full image $\boldsymbol{x}$, the CLIP image encoder outputs the target features $\{\boldsymbol{t}_{j}\}_{j=1}^N=f_{\theta}(\boldsymbol{x})$ where $N$ is the number of image patches. 
We mask a high portion of patches in $\boldsymbol{x}$, and obtain a masked image $\tilde{\boldsymbol{x}}$. The masked autoencoder outputs the reconstructed representations $\{\boldsymbol{p}_{j}\}_{j=1}^N= (h_{\nu}\circ g_{\xi})(\tilde{\boldsymbol{x}})$. We apply $\ell_2$-normalization to both the targets and reconstructions: $\bar{\boldsymbol{t}}_j=\boldsymbol{t}_j/\|\boldsymbol{t}_j\|_2$, $\bar{\boldsymbol{p}}_j=\boldsymbol{p}_j/\|\boldsymbol{p}_j\|_2$ for $j\in[N]$.
Finally, we define the following mean squared error between the normalized target features and reconstructed representations:
\begin{equation}
    \mathcal{L}_{\xi, \nu} = (1/N)\cdot\sum\nolimits_{j=1}^N\|\bar{\boldsymbol{p}}_j-\bar{\boldsymbol{t}}_j\|_2^2. \label{objective}
\end{equation}
MILAN learns the weights $\xi, \nu$ of the masked autoencoder by minimizing objective \ref{objective}.
Note that the reconstruction loss is computed on the features of both masked and unmasked patches.

\subsection{Decoder design: prompting decoder}

Since the decoder is discarded after pretraining, the encoder needs to learn rich semantic information in the latent representations of the unmasked patches from the reconstruction target. To achieve so, 
the functional roles of the encoder and decoder need to be clearly segregated. 
All representation learning for the unmasked patches is completed in the encoder, while the decoder is only for predicting the {target features of the} masked patch{es}.
However, in some previous works~\cite{xie2021simmim,bao2021beit}, the decoder is as simple as one linear layer, which may be insufficient to reconstruct the masked representations, and portions of the encoder may serve as the decoder. MAE \cite{he2021mae} uses a deep decoder that not only updates the mask tokens but also enhances the features of the unmasked patches.
Because our method reconstructs language guided latent representations instead of raw pixels,
the encoder's outputs should only provide clues to the decoder to complete the missing patches' features without being updated in the decoder.
Otherwise, the representation quality from the encoder becomes sub-optimal.



In MILAN, we propose a prompting decoder shown in Figure~\ref{fig:architecture_overview}(b), where the representations of the {unmasked} patches from the encoder {are} frozen, {serving} as ``fixed prompts''. They are appended to the keys and values in each attention module of the prompting decoder's transformer block while the queries only contain the features of mask tokens.
{In specific}, the multi-head attention (MHA) module in Figure \ref{fig:architecture_overview}(b) performs the following operations:
\begin{equation}
{\footnotesize
\begin{split}
\text{MHA}(X, Z) & = \text{Attn}\big( XW_{q},~\text{cat}(Z, XW_{k}),~\text{cat}(Z, XW_{v}) \big), \\
& = \sum_{h=1}^H\text{softmax}(\dfrac{XW_q^h\text{cat}(Z, XW_{k}^h)^T}{\sqrt{d^h}})\text{cat}(Z, XW_{v}^h)W^h_o. \label{decoder_attention}
\end{split}
}
\end{equation}
In \eqref{decoder_attention}, $X$ and $Z$ are the features of mask tokens and the encoder's output{s}, respectively.
``Attn'' is short for the softmax attention operation. ``cat'' means {concatenating} along the sequence dimension. $H$ is the number of heads. $W_q^h, W_k^h, W_v^h, W_o^h$ represent the query, key, value and output projection {weight} matrices in each head. $d^h$ is the embedding dimension in each head.
As shown in Figure \ref{fig:architecture_overview}(c), our {prompting} decoder only computes the self-attention among the features of mask tokens and the cross-attention between the encoder's output and the features of mask tokens.
Moreover, the FFN modules in the prompting decoder only compute on the features of mask tokens. 

Using the default 75\% masking ratio, our prompting decoder reduces the decoding computation cost by 20\% compared to MAE~\cite{he2021mae}. More importantly, prompting decoder improve{s} the finetuning accuracy significantly, as will be analyzed in Section~\ref{sec:classification}.

\subsection{Masking strategy: semantic aware sampling}

To make the masked image pretraining a meaningful pretext task, previous works mask {a} very high portion of input image patches uniformly at random.
With this aggressive masking strategy, it is possible that the remaining few visible patches only contain background information{, which may not provide}
{the} important clues 
{needed to reconstruct} the foreground objects buried in the piles of masked patches, obstructing the model to learn transferable representations.
This becomes a more severe problem in our framework, because the latent representations from the encoder are frozen in the decoding process.
To ensure the representation quality of the pretrained model, previous methods \cite{he2021mae,baevski2022data2vec,wei2021maskfeat} usually require very long pretraining epochs.

To improve the pretraining efficiency, we propose a semantic aware mask sampling strategy
that can make more rational decisions on which patches to mask when using a very high masking ratio. The idea is that the few visible patches fed into the encoder cover important image regions with high probabilities, so that the latent representations from the encoder provide sufficient clues to the decoder to predict the representations of the masked patches.

\begin{table*}[t]
\small
\centering
\begin{tabular}{l l c c c c c}\toprule
    \multirow{2}{*}{\textbf{Method}} & \multirow{2}{*}{\textbf{Training data}} & \multirow{2}{*}{\textbf{Res.}} & \multicolumn{2}{c}{\textbf{ViT-Base}} & \multicolumn{2}{c}{\textbf{ViT-Large}} \\
     & & & Epochs & Top-1 (\%) & Epochs & Top-1 (\%) \\\toprule
     \;\;Supervised \cite{touvron2022deit} & IN1K & 224 & - & 83.8 \textcolor{applegreen}{(+1.6)} & - & 84.9 \textcolor{applegreen}{(+2.9)} \\
     \midrule
     \multicolumn{7}{l}{\textbf{\textit{contrastive or clustering based}}}\\
     \;\;MoCov3 \cite{chen2021mocov3} & IN1K & 224 & 300 & 83.2 \textcolor{applegreen}{(+2.2)} & 300 & 84.1 \textcolor{applegreen}{(+3.7)} \\
     \;\;DINO \cite{caron2021dino} & IN1K & 224 & 400 & 82.8 \textcolor{applegreen}{(+2.6)} & - & - \\
     \;\;Mugs \cite{zhou2022mugs} & IN1K & 224 & 1600 & 84.3 \textcolor{applegreen}{(+1.1)} & - & - \\
     \;\;iBOT \cite{zhou2021ibot} & IN22K\&1K & 224 & 320 & 84.4 \textcolor{applegreen}{(+1.0)} & 200 & 86.3 \textcolor{applegreen}{(+1.5)} \\
     \midrule
     \multicolumn{7}{l}{\textbf{\textit{reconstruction based}}}\\
     \;\;BEiT \cite{bao2021beit} & D250M+IN22K\&1K & 224 & 150 & 83.7 \textcolor{applegreen}{(+1.7)} & 150 & 86.0 \textcolor{applegreen}{(+1.8)} \\
     \;\;mc-BEiT \cite{li2022mcbeit} & OI9M+IN1K & 224 & 800 & 84.1 \textcolor{applegreen}{(+1.3)} & 800 & 85.6 \textcolor{applegreen}{(+2.2)} \\
     \;\;PeCo \cite{dong2021peco} & IN1K & 224 & 800 & 84.5 \textcolor{applegreen}{(+0.9)} & 800 & 86.5 \textcolor{applegreen}{(+1.3)} \\
     \;\;SimMIM \cite{xie2021simmim} & IN1K & 224 & 800 & 83.8 \textcolor{applegreen}{(+1.6)} & - & - \\
     \;\;MaskFeat \cite{wei2021maskfeat} & IN1K & 224 & 1600 & 84.0 \textcolor{applegreen}{(+1.4)} & 1600 & 85.7 \textcolor{applegreen}{(+2.1)} \\
     \;\;data2vec \cite{baevski2022data2vec} & IN1K & 224 & 800 & 84.2 \textcolor{applegreen}{(+1.2)} & 1600 & 86.6 \textcolor{applegreen}{(+1.2)}  \\
     \;\;CAE \cite{chen2022cae} & D250M+IN1K & 224 & 1600 & 83.9 \textcolor{applegreen}{(+1.5)} & 1600 & 86.3 \textcolor{applegreen}{(+1.5)} \\
     \;\;MAE \cite{he2021mae} & IN1K & 224 & 1600 & 83.6 \textcolor{applegreen}{(+1.8)} & 1600 & 85.9 \textcolor{applegreen}{(+1.9)} \\
     \midrule
     \multicolumn{7}{l}{\textbf{\textit{language-image pretraining {based} }}}\\
     \;\;CLIP \cite{radford2021clip} & OpenAI400M+IN1K & 224 & - & 82.1 \textcolor{applegreen}{(+3.3)} & - & 85.3 \textcolor{applegreen}{(+2.5)} \\
    \;\;MVP \cite{wei2022mvp} & OpenAI400M+IN1K & 224 & 300 & 84.4 \textcolor{applegreen}{(+1.0)} & 300 & 86.3 \textcolor{applegreen}{(+1.5)} \\
     \midrule
     {\;\;\textbf{MILAN}} & OpenAI400M+IN1K & 224 & 400 & \textbf{85.4} & 400 & \textbf{87.8} \\
     \bottomrule
     \;\;Supervised \cite{dosovitskiy2020vit} & JFT300M+IN1K & 384 & 90 & 84.2 \textcolor{applegreen}{(+2.2)} & 90 & 87.1 \textcolor{applegreen}{(+1.2)} \\
     \;\;BEiT \cite{bao2021beit} & D250M+IN1K & 384 & 800 & 84.6 \textcolor{applegreen}{(+1.8)} & 800 & 86.3 \textcolor{applegreen}{(+2.0)} \\
     \;\;SWAG \cite{singh2022swag} & IG3.6B+IN1K & 384 & 2 & 85.3 \textcolor{applegreen}{(+1.1)} & - & - \\
     \midrule
     \;\;\textbf{MILAN} & OpenAI400M+IN1K & 384 & 400 & \textbf{86.4} & 400 & \textbf{88.3} \\
     \bottomrule
\end{tabular}
\vspace{-0.1in}
\caption{Comparison of {the} \textbf{finetuning} top-1 accuracy on ImageNet-1K {dataset}.
All models are pretrained with 224$\times$224 input resolution.
We compare finetuning with both 224$\times$224 and 384$\times$384 resolution{s}.
``Epochs'' refer to the pretraining epochs.
``-'': not reported by the original paper. ``IN, D250M, OI9M, IG3.6B'' refer to ImageNet, DALL-E, OpenImages, and Instagram data, respectively.}
\label{tab:classification_finetune}
\end{table*}

To discriminate the semantically important patches {from the} unimportant ones, we use the attention weights from the last self-attention layer in the CLIP image encoder, which
takes the patches of the entire image and an extra class token as input.
Denote the input features to the last self-attention layer of the CLIP image encoder by $[\mathbf{z}_{\text{class}};\mathbf{z}_1;...;\mathbf{z}_{N}]\in\mathbb{R}^{(N+1)\times d}$, where $N$ is the sequence length and $d$ is the embedding dimension.
The interaction between the class token and other features is given by the following attention mechanism:
\begin{equation}
    \mathbf{s}_\text{class} = \text{softmax}({\mathbf{q}_{\text{class}}\mathbf{K}^T}/{\sqrt{d}}), \label{sampling_attention}
\end{equation}
where $\mathbf{s}_{\text{class}}\in\mathbb{R}^{1\times{(1+N)}}$ is the attention vector of the class token. 
$\mathbf{q}_{\text{class}}=\mathbf{z}_{\text{class}}{W}_q$ is the query vector of the class token, and $\mathbf{K}=[\mathbf{z}_{\text{class}};\mathbf{z}_1;...;\mathbf{z}_{N}]W_k$ is the key matrix, where the query and key projection matrices have dimensions $W_q,W_k\in\mathbb{R}^{d\times d}$.
For simplicity, we {show} a single-head attention in \eqref{sampling_attention}. When multiple attention heads {are present}, $\mathbf{s}_\text{class}$ is obtained by averaging over all the heads. 
Because the class token from the last layer of the CLIP image encoder is used to align with the text embedding {from the text encoder}, $\mathbf{s}_\text{class}$ reflect{s} how much information one image patch contributes to the output features of the CLIP image encoder. 
The magnitude of the $i$-th element in $\mathbf{s}_\text{class}$, denoted by $\mathbf{s}_\text{class}{(i)}$, {indicates} whether {the} $i$-th patch is semantically important or not. 
The attention vector $\mathbf{s}_\text{class}$ provides us the premise to design a non-uniform sampling distribution.
Due to the softmax operation, 
we can regard $\mathbf{s}_\text{class}{(i)}$ as the probability of leaving the $i$-th patch unmasked in the input image. Let $r$ represent the masking ratio.
The indices of {the} unmasked patches are obtained by sampling a Multinomial distribution with probabilities $\{\mathbf{s}_\text{class}(0),...,\mathbf{s}_\text{class}(N)\}$
for $\lceil (1-r)N \rceil$ trials without replacement.



\section{Experiments}
We pretrain the ViT-Base and ViT-Large models using MILAN method on ImageNet-1K dataset for 400 epochs using PyTorch framework on 
A100 
machines. The detailed training setup and hyperparameters {can be found} in the appendix. 
We use the CLIP ViT-Base and the CLIP ViT-Large image encoders obtained from OpenAI's paper~\cite{radford2021clip} to produce the reconstruction targets when pretraining our ViT-Base and ViT-Large models, respectively.

\subsection{Classification on ImageNet-1K}
\label{sec:classification}

\noindent\textbf{Finetuning results.}
Table~\ref{tab:classification_finetune} compares the finetuning accuracy on ImageNet-1K {dataset} using MILAN and previous works on the ViT model architecture. 
We pretrain and finetune the ViT models using {ImageNet-1K} dataset only.
Since the CLIP model we use is pretrained on OpenAI's in-house 400M data, we also list it in {the} training data for {MILAN}. However, we only use its image encoder's output features as the reconstruction target for our masked autoencoder.
Even though the supervised ViT models are pretrained on the large {JFT300M} dataset with explicit human labels, MILAN still outperforms them by a clear margin, \textit{e.g.,} improving ViT-Base by +2.2\%.

The self-supervised pretraining methods are divided by using contrastive or reconstruction based objectives. We also compare with large-scale weakly-supervised pretraining using hashtag supervision~\cite{singh2022swag} from an external dataset of 3.6 billion training samples.
MILAN produces higher accuracy than all listed prior arts.
Compared with MAE, MILAN improves the accuracy by +1.8\% for ViT-Base and +1.9\% for ViT-Large.

The CLIP model learns visual representations with language supervision on a large image-text dataset. 
Finetuning the CLIP image encoder
does not lead to competitive accuracy.
However,
when using the image features from the pretrained CLIP model as {the} reconstruction target to train a mask autoencoder, MILAN 
improves the accuracy by 3.3\% on ViT-Base.

\begingroup
\setlength{\tabcolsep}{3pt}
\begin{table}[t]
\centering
\small
\begin{tabular}{l c l c l}\toprule
    \multirow{2}{*}{\textbf{Method}} & \multicolumn{2}{c}{\textbf{ViT-Base}} & \multicolumn{2}{c}{\textbf{ViT-Large}} \\
     & Epochs & \;Top-1 (\%) & Epochs & \;Top-1 (\%) \\\toprule
     \multicolumn{5}{l}{\textbf{\textit{contrastive or clustering based}}}\\
     \;\;MoCov3 \cite{chen2021mocov3} & 300 & 76.7 \textcolor{applegreen}{(+3.2)}& 300 & 77.6 \textcolor{applegreen}{(+6.7)}\\
     \;\;DINO \cite{caron2021dino} & 400 & 78.2 \textcolor{applegreen}{(+1.7)}& - & - \\
     \;\;iBoT \cite{zhou2021ibot} & 1600 & {79.5} \textcolor{applegreen}{(+0.4)}& 1000 & 81.0 \textcolor{applegreen}{(+3.3)}\\
     \midrule
     \multicolumn{5}{l}{\textbf{\textit{reconstruction based}}}\\
     \;\;BEiT \cite{bao2021beit} & 800 & 56.7 \textcolor{applegreen}{(+23.2)}& 800 & 73.5 \textcolor{applegreen}{(+10.8)}\\
     \;\;SimMIM \cite{xie2021simmim} & 800 & 56.7 \textcolor{applegreen}{(+23.2)}& - & - \\
     \;\;MaskFeat \cite{wei2021maskfeat} & - & - & 1600 & 67.7 \textcolor{applegreen}{(+16.6)} \\
     \;\;CAE \cite{chen2022cae} & 1600 & 70.4 \textcolor{applegreen}{(+9.5)} & 1600 & 78.1 \textcolor{applegreen}{(+6.2)} \\
     \;\;MAE \cite{he2021mae} & 1600 & 68.0 \textcolor{applegreen}{(+11.9)}& 1600 & 75.8 \textcolor{applegreen}{(+8.5)}\\
     \midrule
     \multicolumn{5}{l}{\textbf{\textit{language-image pretraining {based}}}}\\
     \;\;CLIP \cite{radford2021clip} & - & 66.5 \textcolor{applegreen}{(+13.4)}& - & 70.5 \textcolor{applegreen}{(+13.8)}\\
    \;\;MVP \cite{wei2022mvp} & 300 & 75.4 \textcolor{applegreen}{(+4.5)} & - & - \\
     \midrule
     \;\;{\textbf{MILAN}} & 400 & \;\;\;\;\textbf{79.9} & 400 & \;\;\;\;\textbf{84.3} \\
     \bottomrule
\end{tabular}
\vspace{-0.1in}
\caption{Comparison of {the} \textbf{linear probing} top-1 accuracy on ImageNet-1K {dataset}. ``Epochs'' refer to the pretraining epochs of various methods. All methods adopt 224$\times$224 input resolution in both pretraining and linear classifier tuning.}
\label{tab:linear_probing}
\end{table}
\endgroup

\noindent\textbf{Linear probing results.}
Instead of finetuning the entire model, we also perform linear probing by appending a linear classifier after the final layer of the pretrained model, and only finetune the linear classifier. Table~\ref{tab:linear_probing} compares the top-1 accuracy on ImageNet-1K {dataset} using various pretraining methods.
As the results show, MILAN beats the accuracy of 
{reconstruction based and language-image pretraining based approaches} by a large margin.
It matches (on ViT-Base) and outperforms (on ViT-Large) previous best contrastive based methods,
which learn more linearly separable representations by instance discrimination \cite{chen2021mocov3} or clustering \cite{caron2021dino}, and are known to be more effective in linear probing \cite{chen2020simclr,he2020moco, chen2021siamese, grill2020byol}.
More linear probing results by other variants of MILAN can be found in the appendix.

\subsection{Downstream tasks}
\label{sec:det_seg}

\noindent\textbf{Object detection and instance segmentation on COCO.}
To verify the transferability of MILAN, we evaluate it on COCO dataset \cite{lin2014coco} for object detection and instance segmentation.
Following MAE \cite{he2021mae}, the pretrained ViT backbones are adapted {to} FPN \cite{lin2017fpn} in the Mask R-CNN framework \cite{he2017maskrcnn}, which is finetuned end-to-end on COCO training set to produce the bounding boxes (evaluated by box AP) and the instance masks (evaluated by mask AP) simultaneously. The results are shown in Table \ref{tab:detseg}.
Compared to supervised pretraining, MILAN performs
better in both tasks, achieving 4.7 and 2.6 points improvements by AP$_{\text{box}}$ and AP$_{\text{mask}}$ on ViT-Base, respectively. 
Compared with {the} previous best result from MAE, which is obtained by 1600-epoch pretraining, 
MILAN advances AP$_{\text{box}}$ and AP$_{\text{mask}}$ by 2.3 and 0.6 points on ViT-Base but {only} pretrains for 400 epochs.

\noindent\textbf{Semantic segmentation on ADE20K.}
We also transfer our pretrained models to semantic segmentation task on the ADE20K dataset \cite{zhou2017ade20k}.
Following the training recipe provided by MAE \cite{he2021mae}, the ViT models pretrained on ImageNet-1K dataset serve as the backbone of UperNet \cite{xiao2018unified}, and are finetuned together with the segmentation layers.
In Table \ref{tab:detseg}, we report the mean intersection over union (mIoU) averaged over all semantic categories.
Our method significantly improves the transferring results of ViT-Base to 52.7, surpassing MAE by 4.6 points.


\begin{table*}[t]
\centering
\small
\begin{tabular}{l c l l l}\toprule
    \multirow{2}{*}{\textbf{Method}} & \multirow{2}{*}{\textbf{Epochs}} & \textbf{Object detection} & \textbf{Instance segmentation} & \textbf{Semantic segmentation} \\
    & & {ViT-B / ViT-L AP$_{\text{box}}$} & {ViT-B / ViT-L  AP$_{\text{mask}}$} & {ViT-B / ViT-L mIoU} \\\toprule
    Supervised \cite{he2017maskrcnn,xiao2018unified} & - & 47.9 \textcolor{applegreen}{(+4.7)} / 49.3 \textcolor{applegreen}{(+6.6)}  & 42.9 \textcolor{applegreen}{(+2.6)} / 43.9 \textcolor{applegreen}{(+4.3)}  & 47.4 \textcolor{applegreen}{(+5.3)} / 49.9 \textcolor{applegreen}{(+8.0)}  \\
    MoCov3 \cite{chen2021mocov3} & 300 & 47.9 \textcolor{applegreen}{(+4.7)} / 49.3 \textcolor{applegreen}{(+6.6)} & 42.7 \textcolor{applegreen}{(+2.8)} / 44.0 \textcolor{applegreen}{(+4.2)} & 47.3 \textcolor{applegreen}{(+5.4)} / 49.1 \textcolor{applegreen}{(+8.8)}  \\
    DINO \cite{caron2021dino} & 300 & 46.8 \textcolor{applegreen}{(+5.8)} / - & 41.5 \textcolor{applegreen}{(+4.0)} / - & 47.2 \textcolor{applegreen}{(+5.5)} / - \\
    BEiT \cite{bao2021beit} & 300 & 42.6 \textcolor{applegreen}{(+10.)} / 53.3 \textcolor{applegreen}{(+2.6)} & 38.8 \textcolor{applegreen}{(+6.7)} / 47.1 \textcolor{applegreen}{(+1.1)} & 45.7 \textcolor{applegreen}{(+7.0)} / 53.3 \textcolor{applegreen}{(+4.6)} \\
    PeCo \cite{dong2021peco} & 300 & 43.9 \textcolor{applegreen}{(+8.7)} / - & 39.8 \textcolor{applegreen}{(+5.7)} / - & 46.7 \textcolor{applegreen}{(+6.0)} / - \\
    SplitMask \cite{el2021splitmask}  & 300 & 46.8 \textcolor{applegreen}{(+5.8)} / - & 42.1 \textcolor{applegreen}{(+3.4)} / - & 45.7 \textcolor{applegreen}{(+7.0)} / - \\
    CAE \cite{chen2022cae} & 1600 & 50.0 \textcolor{applegreen}{(+2.6)} / 54.5 \textcolor{applegreen}{(+1.4)} & 44.0 \textcolor{applegreen}{(+1.5)} / 47.6 \textcolor{applegreen}{(+0.6)} & 50.2 \textcolor{applegreen}{(+2.5)} / 54.7 \textcolor{applegreen}{(+3.2)} \\
    MAE \cite{he2021mae} & 1600 & 50.3 \textcolor{applegreen}{(+2.3)} / 53.3 \textcolor{applegreen}{(+2.6)} & 44.9 \textcolor{applegreen}{(+0.6)} / 47.2 \textcolor{applegreen}{(+1.0)} & 48.1 \textcolor{applegreen}{(+4.6)} / 53.6 \textcolor{applegreen}{(+4.3)} \\
    \midrule
    {\textbf{MILAN}} & 400 & \hspace{0.3in} \textbf{52.6} / \textbf{55.9} & \hspace{0.3in} \textbf{45.5} / \textbf{48.2} & \hspace{0.3in} \textbf{52.7} / \textbf{57.9} \\
    \bottomrule
\end{tabular}
\vspace{-0.1in}
\caption{Results of object detection and instance segmentation are obtained by using Mask R-CNN on COCO with {an} input resolution of 1024$\times$1024. Semantic segmentation results are obtained by using UperNet on ADE20K with {an} input resolution of 512$\times$512.
All methods use ViT models pretrained on ImageNet-1K as backbones. ``Epochs'' refer to the pretraining epochs. ``-'': not reported by the original paper.}
\label{tab:detseg}
\end{table*}

\begin{table*}[t]
\centering
\small
\begin{tabular}{l l l l l }\toprule
    {\textbf{Method}} & {\textbf{Parameters}} & \hspace{0.3in}\textbf{Adversarial} (\%) & \hspace{0.34in}\textbf{Rendition} (\%) & \hspace{0.4in}\textbf{Sketch} (\%) \\\toprule
    Supervised \cite{dosovitskiy2020vit} & 86M / 307M & 27.2  \textcolor{applegreen}{(+35.0)} / 29.6 \textcolor{applegreen}{(+46.3)} & 49.4  \textcolor{applegreen}{(+14.7)} / 50.9 \textcolor{applegreen}{(+25.7)} & 35.6  \textcolor{applegreen}{(+10.7)} / 37.5 \textcolor{applegreen}{(+19.9)} \\
    Swin \cite{liu2021swin} & 88M / - & 35.8  \textcolor{applegreen}{(+26.4)} / - & 46.6  \textcolor{applegreen}{(+17.5)} / - & 32.4  \textcolor{applegreen}{(+13.9)} / - \\
    RobustViT \cite{mao2022robustvit} & 92M / - & 42.3 \textcolor{applegreen}{(+19.9)} / - & 52.6 \textcolor{applegreen}{(+11.5)} / - & 38.4 \textcolor{applegreen}{(+7.9)} ~~/ - \\
    MAE \cite{he2021mae} & 86M / 307M & 35.9 \textcolor{applegreen}{(+26.3)} / 57.1 \textcolor{applegreen}{(+18.8)} & 48.3 \textcolor{applegreen}{(+15.8)} / 59.9 \textcolor{applegreen}{(+16.7)} & 34.5 \textcolor{applegreen}{(+11.8)} / 45.3 \textcolor{applegreen}{(+12.1)} \\
    \midrule
    \textbf{MILAN} & 86M / 307M & \hspace{0.4in}\textbf{62.2 / 75.9} & \hspace{0.4in}\textbf{64.1 / 76.6} & \hspace{0.4in}\textbf{46.3 / 57.4} \\\bottomrule
\end{tabular}
\vspace{-0.1in}
\caption{Comparison of robustness to adversarial examples and distribution shifts on ImageNet datatsets. We evaluate the top-1 accuracy of our MILAN models on different ImageNet validation sets, without any specialized fine-tuning. ``-'': not reported by the original paper.
}
\label{tab:robustness}
\end{table*}

\subsection{Robustness evaluation}
We evaluate the robustness of our models to adversarial examples on ImageNet-Adversarial dataset~\cite{hendrycks2021imagenetadversarial} and distribution shifts on ImageNet-Rendition~\cite{hendrycks2021imagenetrendition} and ImageNet-Sketch~\cite{wang2019imagenetsketch} datasets.
We only finetune our pretrained models on the original ImageNet-1K training set and directly run inference on these different validation sets, without any specialized finetuning.
As shown in Table~\ref{tab:robustness}, MILAN significantly outperforms previous state-of-the-art models. Compared with more advanced architecture RobustViT that is specially designed for robustness and is pretrained on ImageNet-22K, MILAN with vanilla ViT-Base architecture achieves accuracy gains of 7.9\%$\sim$19.9\% on these three datasets. When using ViT-Base, MILAN also surpasses MAE by 26.3\%, 15.8\% and 11.8\% on these three datasets, respectively.

\begingroup
\setlength{\tabcolsep}{2pt}
\begin{table}[t]
\small
\centering
\begin{tabular}{c c c c | l l}\toprule
    & \textbf{CLIP} & \textbf{Prompting} & \textbf{Semantic} & \multirow{2}{*}{\textbf{Epochs}} & \multirow{2}{*}{\textbf{Top-1 (\%)}} \\
    & \textbf{target} & \textbf{decoder} & \textbf{sampling} & & \\\midrule
    \textcolor{gray}{\#1} & \multicolumn{3}{c|}{Baseline (MAE)} & 400 \textcolor{gray}{(1600)} & 83.0 \textcolor{gray}{(83.6)} \\
    \textcolor{gray}{\#2} & \checkmark & & & 400 & 83.9 \\
    \textcolor{gray}{\#3} & & \checkmark & & 400 & 83.0 \\
    \textcolor{gray}{\#4} & & & \checkmark & 400 & 83.3 \\
    \textcolor{gray}{\#5} & & \checkmark & \checkmark & 400 & 83.3 \\
    \textcolor{gray}{\#6} & \checkmark & & \checkmark & 400 & 84.1 \\
    \textcolor{gray}{\#7} & \checkmark & \checkmark & & 400 & 85.1 \\
    \textcolor{gray}{\#8} & \checkmark & \checkmark & \checkmark & 400 \textcolor{gray}{(1600)} & \textbf{85.4} \textcolor{gray}{(85.6)} \\\midrule
    \textcolor{gray}{\#9} & SLIP target & \checkmark & \checkmark & 400 & {84.4} \\
    \bottomrule
\end{tabular}
\vspace{-0.1in}
\caption{
Ablation study of different components in {MILAN}. All results are obtained by pretraining and finetuning ViT-Base {model} on ImageNet-1K {dataset at} 224$\times$224 resolution.
}
\label{tab:ablation}
\vspace{-0.1in}
\end{table}
\endgroup

\subsection{Ablation study}
\label{sec:ablation}

We investigate the effectiveness of {the} different components in 
{MILAN} through {an} ablation {study} in Table \ref{tab:ablation}. More ablation studies on other tasks can be found in the appendix.
Here, the results are based on pretraining {a} ViT-Base model on ImageNet-1K {dataset} for 400 epochs, followed by {a} 100-epoch finetuning. We tune the optimal learning rate for each entry, including the MAE baseline.

(1) By changing the reconstruction target from raw pixels to language guided representations provided by CLIP, the top-1 accuracy is improve{d} by 0.9\% {(\#2 \textit{vs.} \#1 in Table \ref{tab:ablation})}.
{We hypothesize that} 
the CLIP target provides more semantic learning signals for pretraining and encourages the model to get a good grasp of the visual contents instead of {the} low level statistics. 

\begin{figure}[t]
\centering
\includegraphics[scale=0.15]{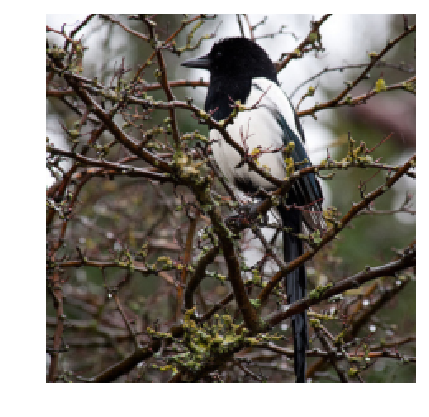}
\includegraphics[scale=0.15]{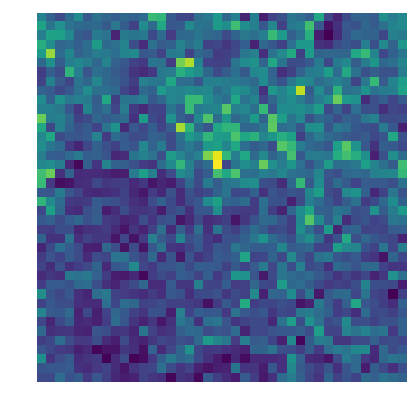}
\includegraphics[scale=0.15]{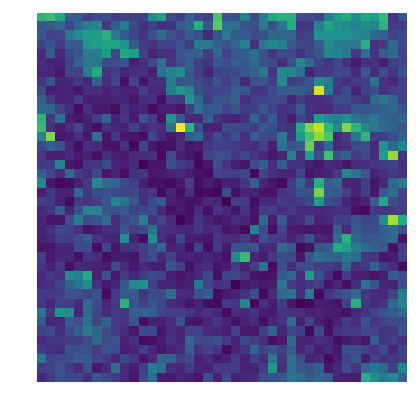}
\includegraphics[scale=0.15]{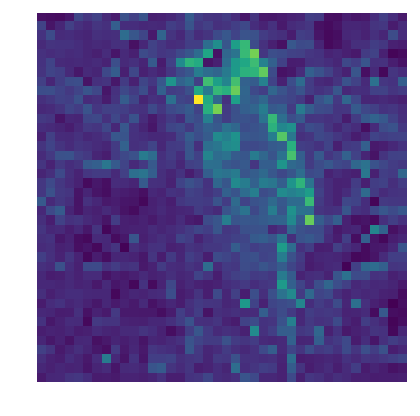}

\includegraphics[scale=0.15]{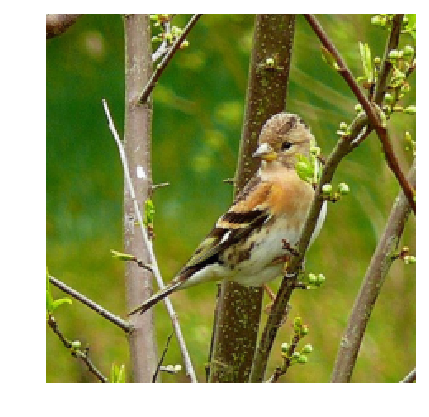}
\includegraphics[scale=0.15]{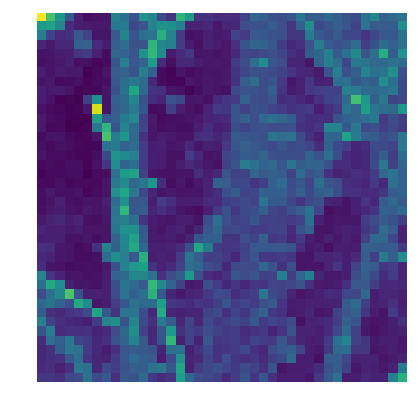}
\includegraphics[scale=0.15]{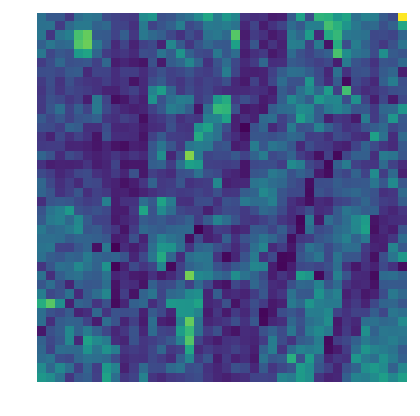}
\includegraphics[scale=0.15]{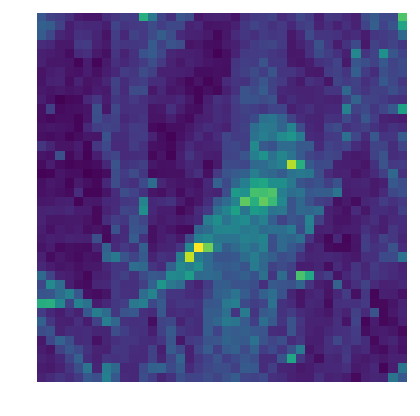}

\includegraphics[scale=0.15]{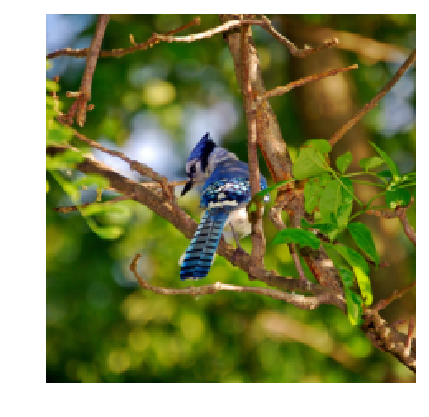}
\includegraphics[scale=0.15]{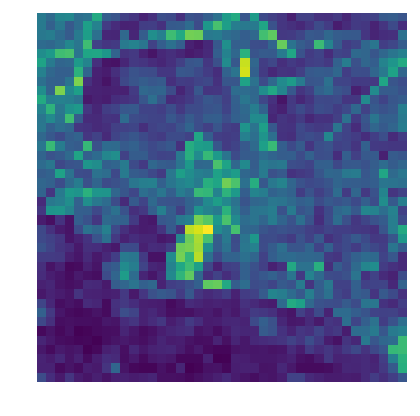}
\includegraphics[scale=0.15]{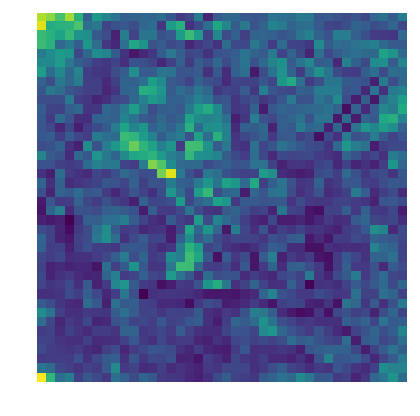}
\includegraphics[scale=0.15]{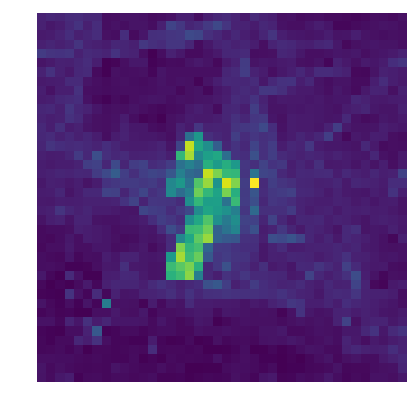}

\includegraphics[scale=0.15]{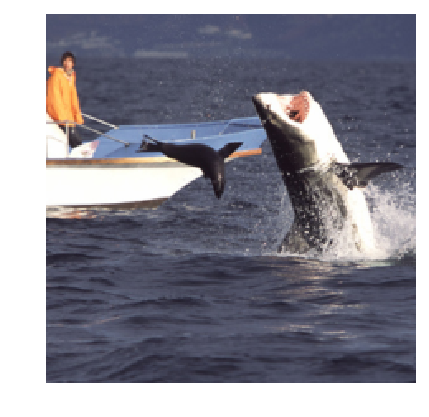}
\includegraphics[scale=0.15]{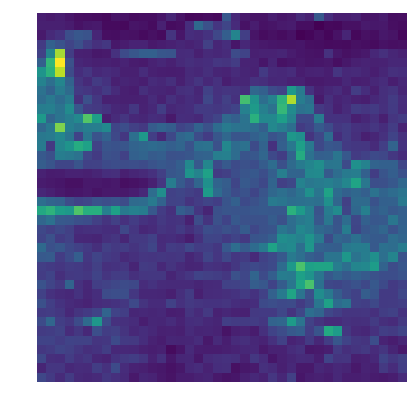}
\includegraphics[scale=0.15]{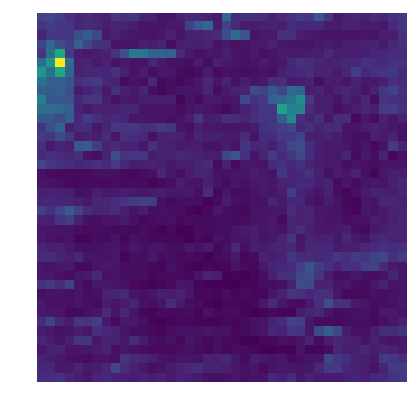}
\includegraphics[scale=0.15]{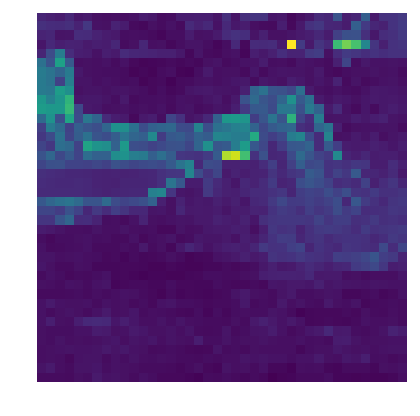}
\vspace{-0.1in}
\centerline{(a) \hspace{0.42in} (b) \hspace{0.42in} (c) \hspace{0.42in} (d)}
\caption{Visualizations of (a) original images and attention features from the last self-attention layer of ViT-Base models pretrained by (b) MAE, (c) MAE+CLIP target, and (d) MILAN.}
\label{fig:ablation_target}
\vspace{-0.2in}
\end{figure}

(2) On top of {the} CLIP target, replacing the original decoder in MAE by our prompting decoder further improves the accuracy by 1.2\% {(\#7 \textit{vs.} \#2 in Table~\ref{tab:ablation})}.
{We also find that the prompting decoder does not increase accuracy when the raw pixels are reconstructed, as the MAE model does (\#3 \textit{vs.} \#1 in Table~\ref{tab:ablation}).}
This big difference can be explained by the different pretraining objectives. When the targets are in the latent space, the encoder's output of {the} unmasked patches' representations should be able to align with the targets without requiring further updates in the decoder. Therefore, in our case of using the CLIP target, the proposed efficient decoder brings in significant accuracy improvements. While in MAE, the encoder's output requires transformation{s} in the decoder to map from {the} latent space {back} to raw pixels.
The results indicate that the decoder design is heavily correlated with the reconstruction target.
We realize the critical coupling effect of these two components and propose mutually beneficial design choices to boost the accuracy.
That means, prompting decoder is specifically designed for the CLIP target and applying it to the pixel target is not beneficial. But when applied to CLIP target, it improves the accuracy significantly.
To further illustrate this insight, we provide visualizations of the learned representations from MAE, MAE+CLIP target, and our MILAN method. As shown in Figure~\ref{fig:ablation_target},
MILAN can better extract the important visual
contents inside the images compared to both MAE and MAE+CLIP target.
This suggests that replacing the reconstruction target alone (MVP \cite{wei2022mvp}) cannot achieve the optimal performance;
the prompting decoder and semantic aware sampling contribute significantly to learning higher quality visual representations on top of the CLIP target.

(3) {The proposed semantic aware mask sampling strategy is generally beneficial regardless of the reconstruction target.} 
After applying the CLIP target and the {prompting} decoder, semantic aware sampling 
further improves {the} uniformally random sampling by 0.3\%, yielding the best 85.4\% accuracy among different model variants {(\#8 \textit{vs.} \#7 in Table~\ref{tab:ablation})}.
Semantic aware sampling slightly reduces the pretraining difficulty and obtains lower training loss, because it favours more important patches.
Although the task becomes easier, this sampling strategy facilitates the model to learn better on objects related regions in the image and
the learned representation enjoys better accuracy.

(4) To demonstrate that 
masked image pretraining generally benefits from the language supervised reconstruction targets, we use a different language-image model, SLIP \cite{mu2021slip}{, to generate the image features as the reconstruction target in MILAN}.
Reconstructing the image features from SLIP still outperforms {reconstructing} raw pixels, surpassing MAE by 1.4\% (\#9 \textit{vs.} \#1 in Table \ref{tab:ablation}).
SLIP incorporates contrastive image self-supervision into language-image pretraining, and finetuning SLIP's image encoder yields an {accuracy} of 82.6\%.
Similar to our observation on CLIP, one more step of reconstruction based pretraining by MILAN further improves the representation quality, boosting the accuracy by 1.8\% compared to finetuning SLIP.

Finally, we note that {a} very long pretraing schedule is no longer necessary for {MILAN} compared to MAE. Our method enjoys much fewer epochs while achieving higher accuracy: {MILAN} achieves 85.4\% after a 400-epoch pretraining, while MAE {achieves} 83.6\% after a 1600-epoch pretraining.





\section{Related Works}

\noindent\textbf{Masked image pretraining.}
Self-supervised pretraining aims to learn transferable representations from unlabeled data by a pretext task \cite{caron2018deepcluster, doersch2015contextpred, van2018cpc, ermolov2021whitening, li2020pcl, goyal2021selfwild, zbontar2021barlowtwins}.
Recent pretraining methods revitalize the use of denoising autoencoders \cite{vincent2010denoising, pathak2016contextencoder, chen2020igpt, zhang2016colorful, noroozi2016jigsaw} to train the vision transformer with masked prediction objectives.
The model receives incomplete {images} with {a large portion of the} patches removed and learns to reconstruct the missing contents {pixel by pixel}~\cite{he2021mae,xie2021simmim, li2021mst}.
To inject semantics into {the} representations, BEiT \cite{bao2021beit}, PeCo \cite{dong2021peco} and CAE \cite{chen2022cae} predict discrete visual vocabularies produced from separately trained tokenizer{s} \cite{rolfe2016dvae, van2017vqvae, ramesh2021dalle}.
MaskFeat~\cite{wei2021maskfeat}
find{s} that {the} local gradient features produced by the manually-crafted HOG descriptor surpasses more complex targets.
Other works like iBOT \cite{zhou2021ibot}, data2vec \cite{baevski2022data2vec}, and \cite{kakogeorgiou2022attentionmask} adopts self-distillation, where the model reconstructs the masked patches' representations produced by the exponential moving average of the model.
{MILAN} differs that it learns the masked autoencoder by reconstructing {the} latent representations that embed rich semantic information stemming from language supervision. 
Moreover, we 
find {that} freezing the encoder's output features in the decoder is a critical factor when the model learns to reconstruct {the} latent targets.
Finally, 
we propose {a} semantic aware mask sampling {mechanism} and alleviate the need for very long pretraining.

\noindent\textbf{Language-Image pretraining.}
Learning visual representations from language supervision is not new.
Early work \cite{frome2013devise}
embeds images and texts into a shared semantic space so that the model is able to recognize classes even without explicit labels. 
Other methods leverage the caption supervision to train the vision model by completing the image captioning task \cite{desai2021virtex} or {the} masked language modeling task \cite{sariyildiz2020icmlm}.
Recently, benefiting from contrastive training \cite{yuan2021multimodal} and {the} scalability of modern backbones, CLIP \cite{radford2021clip} and ALIGN \cite{jia2021align} learn strong visual representations on large-scale image-text dataset{s}, advancing the transfer performance on {the} downstream vision tasks.
{Later} works improve CLIP by introducing more {auxiliary} loss functions {to assist the image-text contrastive loss}, such as image self-supervision loss \cite{mu2021slip, li2021declip}, self-distillation loss \cite{cheng2021data}, and token-wise max similarity \cite{yao2021filip}.
In this work,
MILAN further improve{s} the representation quality of language-image pretraining by incorporating masked image pretraining.

\noindent\textbf{Contrastive learning.}
Contrastive methods \cite{caron2021dino, chen2021mocov3, chen2020simclr, he2020moco, tian2020goodview, chen2021siamese, grill2020byol, caron2020clusterassignment, wu2018nonparametricinstance, wang2021densecontrastive, xie2021pixelconsistency, alexey2016discriminative} learn augmentation invariance by enforcing similarity between {different} views augmented from the same image while avoiding model collapse.
The learned representations show high linear separability and are commonly evaluated by linear probing.
However, contrastive learning heavily depends on strong data augmentations and effective negative sampling.
In contrast, MILAN uses a masked prediction objective with a reconstruction loss. Our method learns powerful representations with much simpler data augmentations.

\section{Conclusion}
The masked autoencoders can extract visual concepts from the unlabeled raw image pixels, which reduces the heavy reliance on large labeled datasets in computer vision tasks. 
However, such visual concepts may still lag the rich semantic data the image captions contain. 
Understanding images assisted by language captions has also been explored in the weakly supervised pretraining setting. The learned features may easily be transferred to downstream tasks via zero shot learning. However, finetuning those models directly  may not reveal competitive results. 
In this paper, we combined these two lines of work to use the outputs of language-image pretraining as the reconstruction target for  masked autoencoders, proposed a more effective decoder architecture and {a semantic aware} sampling mechanism. We have shown that by combining the two methods in the self-supervised pretraining, we can achieve better quality than applying each method individually.

{\small
\bibliographystyle{ieee_fullname}
\bibliography{egbib}
}

\newpage

\appendix
\section{Appendix}

\subsection{Implementation details}

\paragraph{Pretraining on ImageNet-1K.}
We follow the pretraining recipe of MAE \cite{he2021mae}, using a publicly released codebase~\footnote{\url{https://github.com/facebookresearch/mae}}. We only pretrain ViT models for 400 epochs. We use AdamW optimizer with a momentum of $(\beta_1=0.9, \beta_2=0.95)$, a mini-batch size of 4096 and an initial learning rate of $2.4e-3$ (scaled based on $\text{lr}=\text{base\_lr} \times \text{batchsize} / 256$ where $\text{base\_lr}=1.5e-4$). The learning rate is linearly warmed up for the first 40 epochs, and decayed to zero by a cosine learning rate schedule. The weight decay is set to 0.05. For data augmentation, we only adopt random resized crop to $224\times224$ resolution, random horizontal flip, and normalization.
The masking ratio is set to 75\%.
We use {the image features generated from}
pretrained CLIP models with ViT image encoders {as the reconstruction targets}.
{The wall-clock pretraining time of MILAN for ViT-Base and ViT-Large models take} 2 days and 3 days on a machine with 8 A100 GPUs, respectively.
For reference, MAE requires 1600-epoch pretraining which takes 116 hours on ViT-Base model.

\paragraph{Finetuning on ImageNet-1K.}
We follow the finetuning recipe of MAE \cite{he2021mae} but tune the optimal learning rates for our models. We only finetune 100 epochs. We use AdamW optimizer with a momentum of $(\beta_1=0.9, \beta_2=0.999)$, a mini-batch size of 1024 and an initial learning rate of $4e-4$ (scaled based on $\text{lr}=\text{base\_lr} \times \text{batchsize} / 256$ where $\text{base\_lr}=1e-4$). The learning rate is linearly warmed up for the first 5 epochs, and decayed to zero by a cosine learning rate schedule. The layer-wise learning rate decay factor is set to 0.65 for ViT-Base and 0.75 for ViT-Large. The weight decay is set to 0.05. We adopt RandAugment \cite{cubuk2020randaugment}, and set $\text{label\_smoothing}=0.1, \text{mixup}=0.8, \text{cutmix}=1.0, \text{drop\_path}=0.1 (\text{ViT-Base}), 0.2 (\text{ViT-Large})$.

\paragraph{Linear probing on ImageNet-1K.}
We follow the linear probing recipe of MAE \cite{he2021mae}. 
We train the linear classifier for 100 epochs.
We use LARS \cite{you2017lars} optimizer with a momentum of 0.9, a mini-batch size of 16384, and an initial learning rate of 3.2 for ViT-Base and 1.28 for ViT-Large.
The learning rate is linearly warmed up for the first 10 epochs, and decayed to zero by a cosine learning rate schedule.
We do not use mixup, cutmix, drop path, or color jittering, and the weight decay is set to zero.

\paragraph{Objection detection and instance segmentation on COCO.}
Following \cite{he2021mae}, the pretrained ViT models by MILAN are adapted to FPN \cite{lin2017fpn} in the Mask R-CNN framework \cite{he2017maskrcnn}, and we finetuned end-to-end on COCO training set \cite{lin2014coco}.
We use AdamW optimizer with a momentum of $(\beta_1=0.9, \beta_2=0.999)$, a mini-batch size of 16, and an initial learning rate of $2e-4$.
The layer-wise learning rate decay is set to 0.75 for ViT-Base and 0.85 for ViT-Large. The weight decay is set to 0.1 and the drop path rate is set to 0.1 for ViT-Base and 0.2 for ViT-Large. We adopt the standard $1\times$ schedule: 12 epochs with the learning rate decayed by 10 at epochs 8 and 11. The input resolution is $1024 \times 1024$. We do not use multi-scale testing.  We build upon a public codebase~\footnote{\url{https://github.com/hustvl/MIMDet}} that reproduces MAE's detection results.

\paragraph{Semantic segmentation on ADE20K.}
Following \cite{he2021mae}, the ViT models pretrained on ImageNet-1K dataset by MILAN serve as the backbone of UperNet framework\cite{xiao2018unified}, and are finetuned together with the segmentation layers on ADE20K dataset \cite{zhou2017ade20k} for 160K iterations.
We use AdamW optimizer with a momentum of $(\beta_1=0.9, \beta_2=0.999)$, a mini-batch size of 16 and an initial learning rate of $3e-5$. The learning rate is linearly warmed up for the first 1500 iterations, and decayed to zero by a poly learning rate schedule. The layer-wise learning rate decay is set to 0.9. The weight decay is set to 0.05 and the drop path rate is set to 0.1. The input resolution is $512 \times 512$. We do not use multi-scale testing. We build upon a public codebase~\footnote{\url{https://github.com/implus/mae_segmentation}} that reproduces MAE's segmentation results.


\subsection{More results}

\begin{table*}[t]
\small
\centering
\begin{tabular}{l l c c}\toprule
     & \textbf{Method} & \textbf{mIoU} \\\toprule
    \textcolor{gray}{\#1} & Baseline (MAE) & 48.1 \\
    \textcolor{gray}{\#2} & \; + CLIP target & 49.2 \\
    \textcolor{gray}{\#3} & \; + CLIP target + Semantic aware sampling & 50.4 \\
    \textcolor{gray}{\#4} & \; + CLIP target + Prompting  decoder & 51.9 \\
    \textcolor{gray}{\#5} & \; + CLIP target + Prompting  decoder + Semantic aware sampling & \textbf{52.7} \\
    \bottomrule
\end{tabular}
\caption{
Ablation study on different components of {MILAN} on the semantic segmentation task. All results are based on ViT-Base model{s} that are pretrained on ImageNet-1K and finetuned on the ADE20K dataset using the UperNet segmentation framework.
}
\label{tab:semantic_segmentation_ablation}
\end{table*}

\paragraph{Ablation study on semantic segmentation task.}
We also conduct an ablation study on the different components of MILAN on the semantic segmentation task, as shown in Table~\ref{tab:semantic_segmentation_ablation}.
The results are based on pretraining a ViT-Base model on ImageNet-1K dataset for 400 epochs, followed by finetuning UperNet on ADE20K by 160K iterations.
We find that the overall trend is consistent with our findings in the finetuning results on ImageNet.
By changing the reconstruction target from raw pixels to image features produced from CLIP, the mIoU is improved by 1.1 points (\#2 \textit{vs.} \#1 in Table~\ref{tab:semantic_segmentation_ablation}).
On top of the CLIP target, replacing the random masking in MAE by our semantic aware sampling (\#3 \textit{vs.} \#2 in Table~\ref{tab:semantic_segmentation_ablation}) or replacing the vanilla decoder in MAE by our prompting decoder (\#4 \textit{vs.} \#2 in Table~\ref{tab:semantic_segmentation_ablation}) further improves the mIoU by 1.2 and 2.7 points, respectively.
Finally, applying both the semantic aware sampling and the prompting decoder leads to the best 52.7 mIoU, which is 3.5 points higher than applying the CLIP target alone and
4.6 points higher than the MAE baseline. These results provide extra evidence that changing the reconstruction targets alone (e.g., MVP \cite{wei2022mvp}) cannot achieve the optimal performance. The majority of the accuracy gains come from more effective prompting decoder and semantic aware sampling in our method.

\begin{table*}[t]
\small
\centering
\begin{tabular}{l l c}\toprule
     & \textbf{Method} & \textbf{Top-1 (\%)} \\\toprule
    \textcolor{gray}{\#1} & Baseline (MAE) & 62.0 \\
    \textcolor{gray}{\#2} & \; + CLIP target & 67.1 \\
    \textcolor{gray}{\#3} & \; + CLIP target + Semantic aware sampling & 68.1 \\
    \textcolor{gray}{\#4} & \; + CLIP target + Prompting  decoder & {79.9} \\
    \textcolor{gray}{\#5} & \; + CLIP target + Prompting  decoder + Semantic aware sampling & 78.9 \\\midrule
    \textcolor{gray}{\#6} & \; \#5 + Semantic aware probing & \textbf{80.0} \\
    \bottomrule
\end{tabular}
\caption{
Ablation study on different components of {MILAN} on the {linear probing} task.
All results are based on ViT-Base model{s} that are pretrained on ImageNet-1K {dataset at} 224$\times$224 resolution.
}
\label{tab:linear_probing_ablation}
\end{table*}

\begin{table*}[t]
\centering
\small
\begin{tabular}{l c l c l}\toprule
    \multirow{2}{*}{\textbf{Method}} & \multicolumn{2}{c}{\textbf{ViT-Base}} & \multicolumn{2}{c}{\textbf{ViT-Large}} \\
     & Epochs & \;Top-1 (\%) & Epochs & \;Top-1 (\%) \\\toprule
     \multicolumn{5}{l}{\textbf{\textit{contrastive or clustering based}}}\\
     \;\;MoCov3 \cite{chen2021mocov3} & 300 & 76.7 \textcolor{applegreen}{(+3.2)}& 300 & 77.6 \textcolor{applegreen}{(+6.7)}\\
     \;\;DINO \cite{caron2021dino} & 400 & 78.2 \textcolor{applegreen}{(+1.7)}& - & - \\
     \;\;iBoT \cite{zhou2021ibot} & 1600 & {79.5} \textcolor{applegreen}{(+0.4)}& 1000 & 81.0 \textcolor{applegreen}{(+3.3)}\\
     \midrule
     \multicolumn{5}{l}{\textbf{\textit{reconstruction based}}}\\
     \;\;BEiT \cite{bao2021beit} & 800 & 56.7 \textcolor{applegreen}{(+23.2)}& 800 & 73.5 \textcolor{applegreen}{(+10.8)}\\
     \;\;SimMIM \cite{xie2021simmim} & 800 & 56.7 \textcolor{applegreen}{(+23.2)}& - & - \\
     \;\;MaskFeat \cite{wei2021maskfeat} & - & - & 1600 & 67.7 \textcolor{applegreen}{(+16.6)} \\
     \;\;CAE \cite{chen2022cae} & 800 & 68.3 \textcolor{applegreen}{(+11.6)}& - & - \\
     \;\;MAE \cite{he2021mae} & 1600 & 68.0 \textcolor{applegreen}{(+11.9)}& 1600 & 75.8 \textcolor{applegreen}{(+8.5)}\\
     \midrule
     \multicolumn{5}{l}{\textbf{\textit{language-image pretraining {based}}}}\\
     \;\;CLIP \cite{radford2021clip} & - & 66.5 \textcolor{applegreen}{(+13.4)}& - & 70.5 \textcolor{applegreen}{(+13.8)}\\
    \;\;MVP \cite{wei2022mvp} & 300 & 75.4 \textcolor{applegreen}{(+4.5)} & - & - \\
     \midrule
     \;\;{\textbf{MILAN w/ SAS}} & 400 & 78.9 \textcolor{applegreen}{(+1.0)} & 400 & 84.1 \textcolor{applegreen}{(+0.2)} \\
     \;\;{\textbf{MILAN w/ RS}} & 400 & \;\;\;\;\textbf{79.9} & 400 & \;\;\;\;\textbf{84.3} \\
     \bottomrule
\end{tabular}
\caption{Comparison of {the} {linear probing} top-1 accuracy on ImageNet-1K {dataset}. ``SAS'': semantic aware sampling. ``RS'': random sampling. ``Epochs'' refer to the pretraining epochs of various methods. All methods adopt 224$\times$224 input resolution in pretraining and linear probing.}
\label{tab:linear_probing_appendix}
\end{table*}

\paragraph{Ablation study on linear probing task.}
We further conduct {an} ablation study on {the different} components of MILAN on linear probing task, as shown in Table \ref{tab:linear_probing_ablation}. 
The results are based on pretraining a ViT-Base model on ImageNet-1K dataset for 400 epochs, followed by a 100-epoch linear classifier training. 
Consistent with our findings on finetuning and semantic segmentation tasks, using CLIP target brings in accuracy improvement.
By changing the reconstruction target from raw pixels to the semantic preserving CLIP features, the top-1 accuracy {is} boost{ed} by 5\% (\#2 \textit{vs.} \#1 in Table \ref{tab:linear_probing_ablation}).
On top of the CLIP target, replacing the random masking in MAE by our semantic aware mask sampling (\#3 \textit{vs.} \#2 in Table \ref{tab:linear_probing_ablation}) or replacing the original decoder in MAE by our prompting decoder (\#4 \textit{vs.} \#2 in Table \ref{tab:linear_probing_ablation}) improves the accuracy by 1\% and 12.8\%, respectively.
However, inconsistent with our findings on finetuning and semantic segmentation tasks, applying both the prompting decoder and the semantic aware mask sampling simultaneously does not lead to the best accuracy (\#5 \textit{vs.} \#4 in Table \ref{tab:linear_probing_ablation}).
We hypothesize that the encoder model is overly adapted to those unmasked important image patches when semantic aware sampling and prompting decoder are applied together.
In linear probing, the pretrained encoder model receives full patches, and its weights are frozen. Only the linear classifier's weights are updated. 
Thus, the model may not be able to cope with the image patches that are less relevant or totally irrelevant to the objects that need to be classified, and those features degrade the performance of the linear classifier.

To {support our speculation} that the pretrained encoder may be biased towards important patches,
we perform a semantic aware probing experiment, listed as row \#6 in Table \ref{tab:linear_probing_ablation}.
Specifically, the pretrained model is obtained by MILAN with CLIP target, prompting decoder and semantic aware sampling. 
When finetuning the linear classifier as well as performing inference on the validation dataset, we select the top 50\% important image patches. Only the selected patches are fed into the frozen encoder model to obtain the features to train the linear classifier.
Although the classifier is trained on features from incomplete inputs, semantic aware probing improves the accuracy to 80\% (\#6 \textit{vs.} \#5 in Table \ref{tab:linear_probing_ablation}), indicating that the pretrained encoder model is more adept at extracting features from semantically important patches.

From our ablation study on the linear probing task, we find that MILAN with random masking gives better accuracy the ViT-Base model. In Table \ref{tab:linear_probing_appendix}, we include linear probing results obtained by applying MILAN with random masking on both ViT-Base and ViT-Large models.
Compared with MILAN with semantic aware sampling, the linear probing accuracies are further improved by 1\% and 0.2\%, respectively.
Compared with the state-of-the-art contrastive method \cite{zhou2021ibot}, MILAN with random sampling achieves higher linear probing accuracy on both ViT-Base (+0.4\%) and ViT-Large (+3.3\%) models.


\begin{table*}[t]
\centering
\small
\begin{tabular}{l | c c c c c}\toprule
    \multirow{2}{*}{CLIP image encoder} & FT on IN1K & LP on IN1K & OD on COCO & IS on COCO & SS on ADE20K \\
    & top-1 (\%) & top-1 (\%) & AP$_{\text{box}}$ & AP$_{\text{mask}}$ & mIoU \\\midrule
    ViT-Base & 86.7 & 81.6 & 55.0 & 47.5 & 55.3 \\
    ViT-Large & 87.8 & 84.3 & 55.9 & 48.2 & 57.9 \\\bottomrule
\end{tabular}
\caption{Comparison of using different CLIP image encoders to produce the reconstruction targets for pretraining ViT-Large. We compare the results of finetuning (FT) and linear probing (LP) on ImageNet-1K (IN1K), object detection (OD) and instance segmentation (IS) on COCO, and semantic segmentation (SS) on ADE20K. 
}
\label{tab:image_encoder}
\end{table*}

\begin{table}[t]
\centering
\small
\begin{tabular}{l c c}\toprule
    \textbf{Method} & \textbf{Epochs} & \textbf{Top-1 (\%)} \\\midrule
    \;KD & 400 & 84.0 \\
    \;MILAN & 400 & 85.4 \\
    \bottomrule
\end{tabular}
\caption{Compare MILAN with knowledge distillation (KD) on ImageNet-1K using ViT-Base model. Both methods use the representations produced by the CLIP image encoder as the target features. We report the finetuning top-1 accuracy. ``Epoch'' refers to the pretraining epochs.}
\label{tab:milan_vs_kd}
\end{table}

\paragraph{MILAN \textit{vs.} knowledge distillation.}
In MILAN, the decoder reconstructs the representations of the masked patches with the assistance from the encoder's output features of the unmasked patches. {T}he reconstruction loss is computed on both the encoder's output features of the unmasked patches and decoder's output features of the masked patches.
Here, we perform another ablation by removing the decoder from MILAN. 
The pretraining objective becomes training the encoder only to predict the target features on the unmasked patches, which can be regarded as a semantic aware knowledge distillation (KD) method.
For KD, we also use the image features from the pretrained CLIP image encoder as the target, and the pretraining loss is the mean squared error between the normalized target features and the encoder's output features. After pretraining, the encoder model is finetuned end-to-end with the same recipe a MILAN.
The results are shown in Table \ref{tab:milan_vs_kd}.
KD achieves 84.0\% top-1 accuracy, while direct finetuning of the CLIP image encoder only yields 82.1\% accuracy. Due to the possible data distribution gap between the pretraining data (OpenAI's 400M) and the finetuning data (ImageNet-1K), finetuning the CLIP image encoder may not lead to a strong performance. 
But the gap can be overcome by pretraining the model with a KD objective on ImageNet-1K followed by finetuning.
Moreover, MILAN achieves 1.4\% higher accuracy than KD.
MILAN creates a more challenging pretraining task, where the model not only needs to predict the latent features of the visible unmasked patches but also learns to reconstruct the representations of the invisible masked patches through the decoder. The masked image reconstruction task can better leverage the guidance from the target features and improve the representation quality.

\begin{figure*}[t]
\centering
\includegraphics[scale=0.15]{compare_attn/raw/img_907.png}
\hspace{-0.1in}
\includegraphics[scale=0.15]{compare_attn/mae/img_907.png}
\hspace{-0.1in}
\includegraphics[scale=0.15]{compare_attn/ablation/img_907.png}
\hspace{-0.1in}
\includegraphics[scale=0.15]{compare_attn/milan/img_907.png}
\hspace{0.1in}
\includegraphics[scale=0.15]{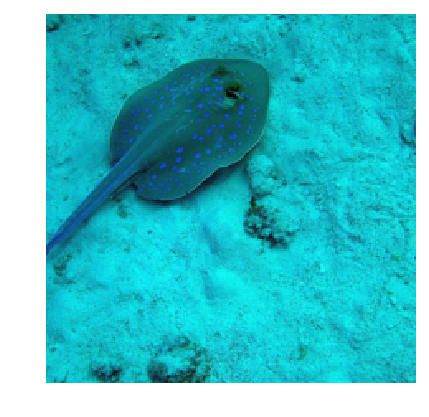}
\hspace{-0.1in}
\includegraphics[scale=0.15]{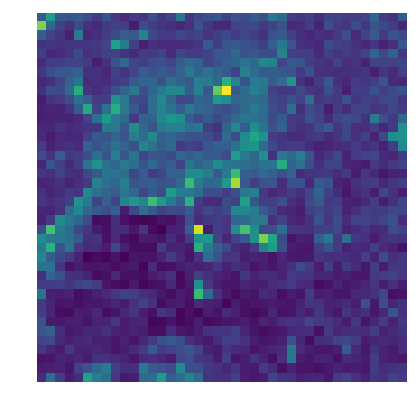}
\hspace{-0.1in}
\includegraphics[scale=0.15]{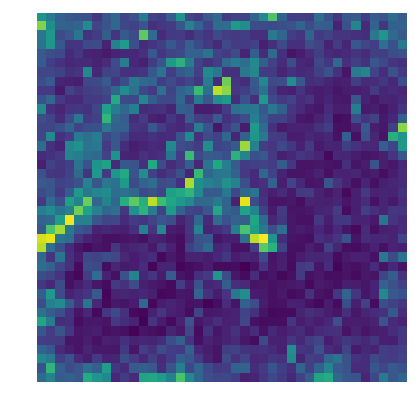}
\hspace{-0.1in}
\includegraphics[scale=0.15]{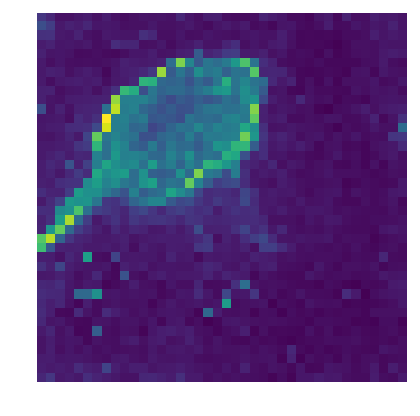}

\includegraphics[scale=0.15]{compare_attn/raw/img_507.png}
\hspace{-0.1in}
\includegraphics[scale=0.15]{compare_attn/mae/img_507.png}
\hspace{-0.1in}
\includegraphics[scale=0.15]{compare_attn/ablation/img_507.png}
\hspace{-0.1in}
\includegraphics[scale=0.15]{compare_attn/milan/img_507.png}
\hspace{0.1in}
\includegraphics[scale=0.15]{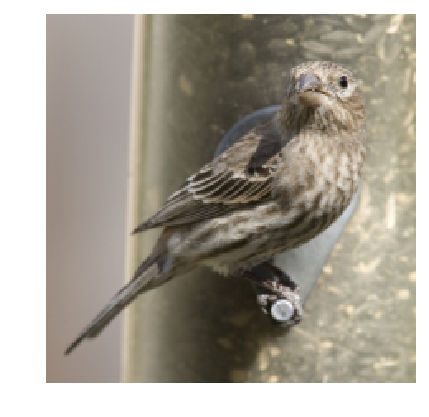}
\hspace{-0.1in}
\includegraphics[scale=0.15]{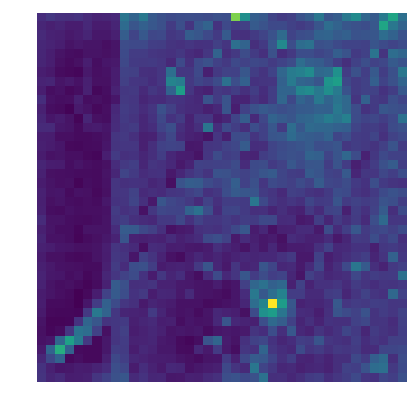}
\hspace{-0.1in}
\includegraphics[scale=0.15]{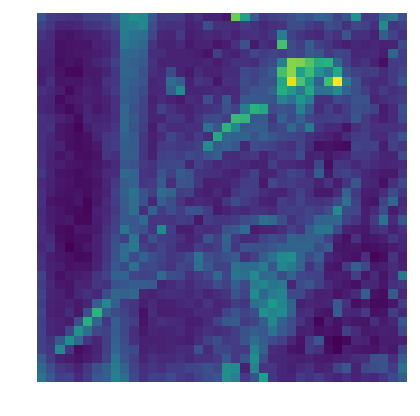}
\hspace{-0.1in}
\includegraphics[scale=0.15]{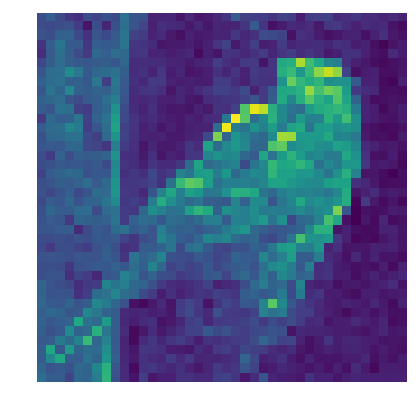}

\includegraphics[scale=0.15]{compare_attn/raw/img_850.png}
\hspace{-0.1in}
\includegraphics[scale=0.15]{compare_attn/mae/img_850.png}
\hspace{-0.1in}
\includegraphics[scale=0.15]{compare_attn/ablation/img_850.png}
\hspace{-0.1in}
\includegraphics[scale=0.15]{compare_attn/milan/img_850.png}
\hspace{0.1in}
\includegraphics[scale=0.15]{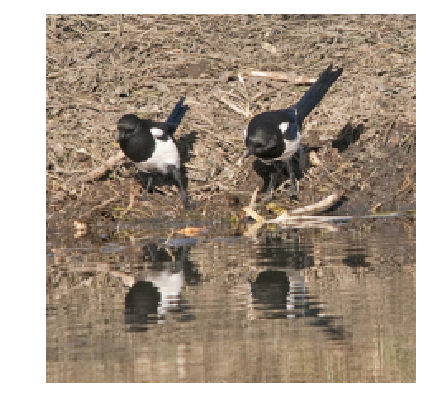}
\hspace{-0.1in}
\includegraphics[scale=0.15]{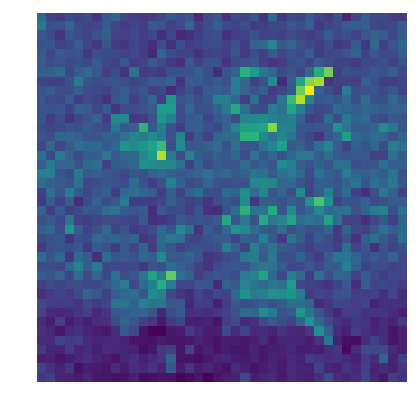}
\hspace{-0.1in}
\includegraphics[scale=0.15]{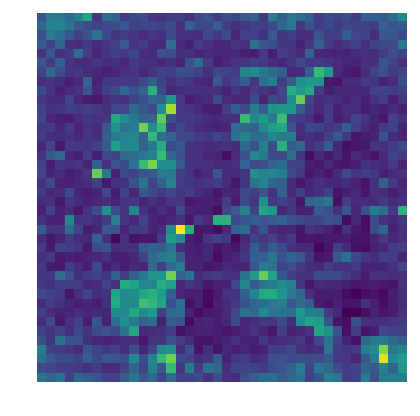}
\hspace{-0.1in}
\includegraphics[scale=0.15]{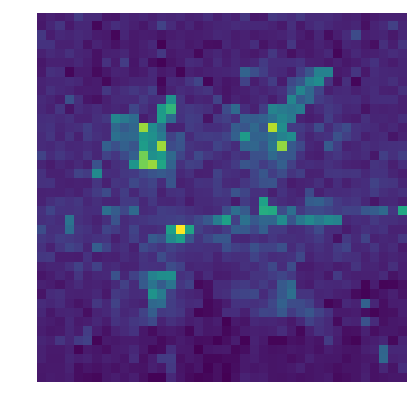}

\includegraphics[scale=0.15]{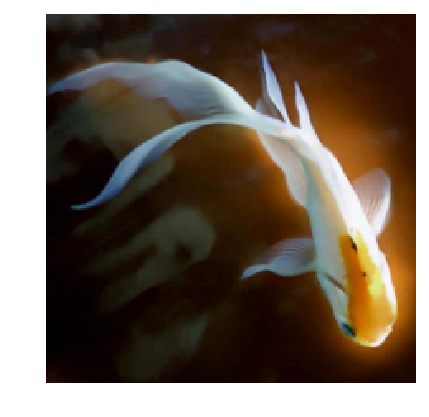}
\hspace{-0.1in}
\includegraphics[scale=0.15]{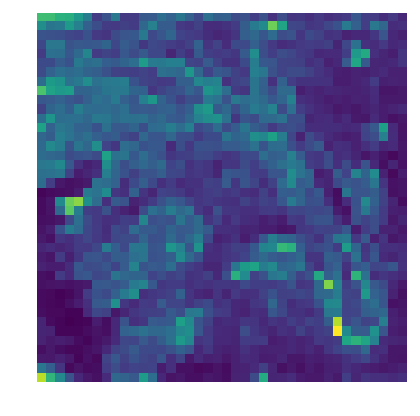}
\hspace{-0.1in}
\includegraphics[scale=0.15]{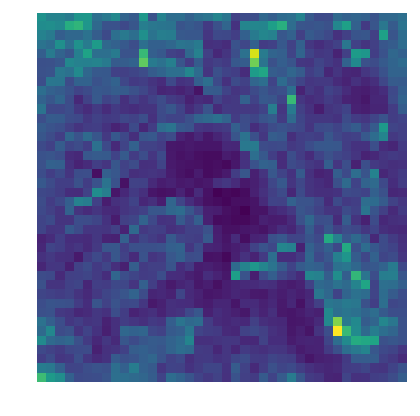}
\hspace{-0.1in}
\includegraphics[scale=0.15]{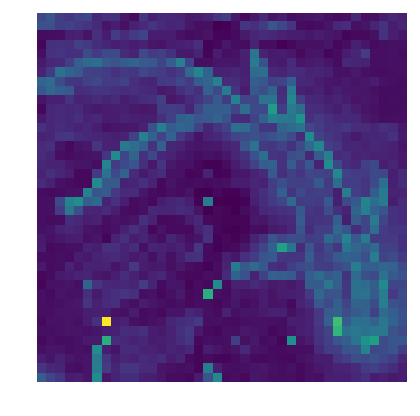}
\hspace{0.1in}
\includegraphics[scale=0.15]{compare_attn/raw/img_107.png}
\hspace{-0.1in}
\includegraphics[scale=0.15]{compare_attn/mae/img_107.png}
\hspace{-0.1in}
\includegraphics[scale=0.15]{compare_attn/ablation/img_107.png}
\hspace{-0.1in}
\includegraphics[scale=0.15]{compare_attn/milan/img_107.png}

\includegraphics[scale=0.15]{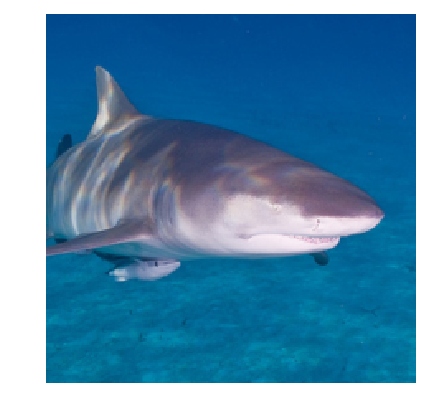}
\hspace{-0.1in}
\includegraphics[scale=0.15]{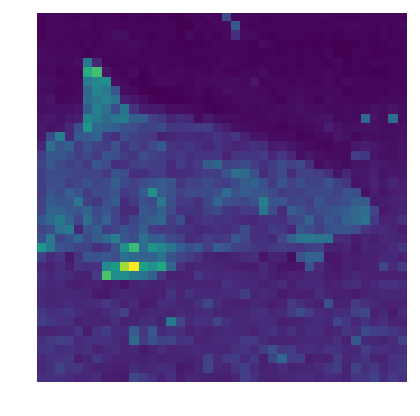}
\hspace{-0.1in}
\includegraphics[scale=0.15]{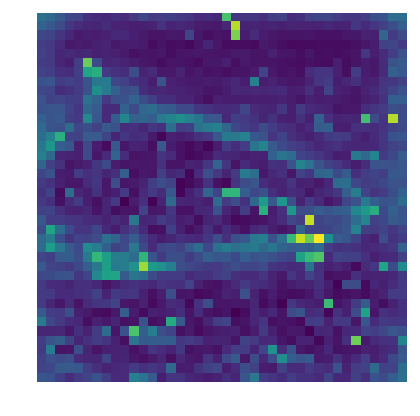}
\hspace{-0.1in}
\includegraphics[scale=0.15]{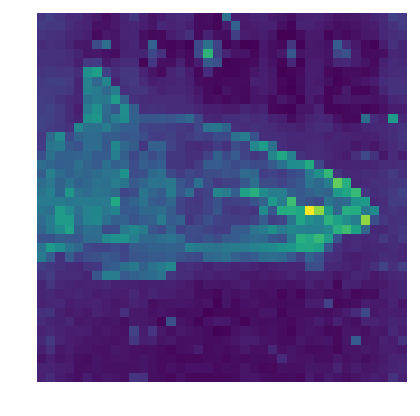}
\hspace{0.1in}
\includegraphics[scale=0.15]{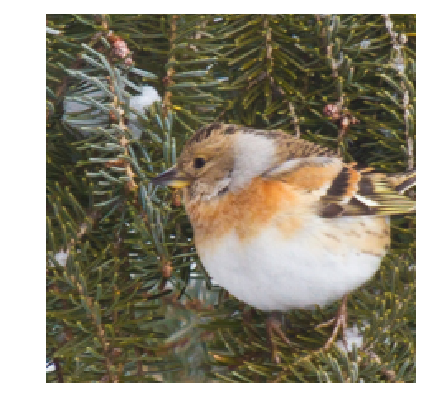}
\hspace{-0.1in}
\includegraphics[scale=0.15]{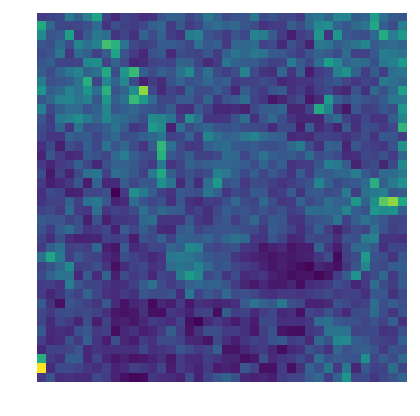}
\hspace{-0.1in}
\includegraphics[scale=0.15]{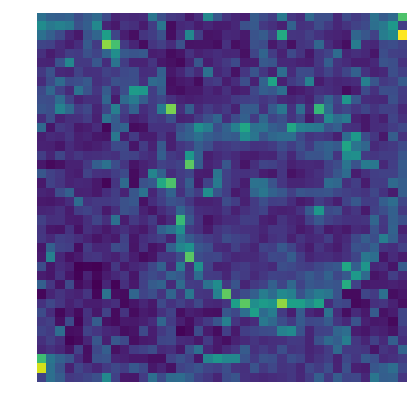}
\hspace{-0.1in}
\includegraphics[scale=0.15]{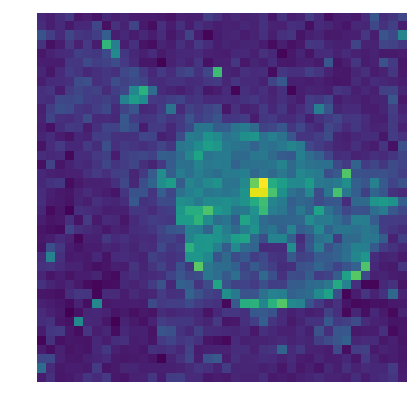}

\includegraphics[scale=0.15]{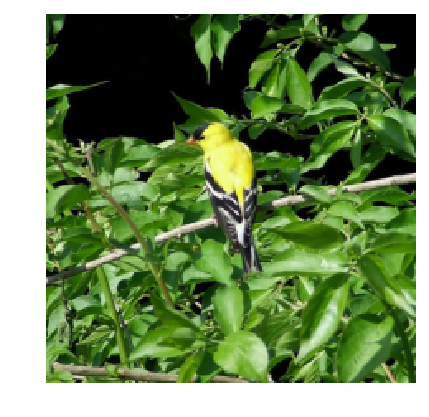}
\hspace{-0.1in}
\includegraphics[scale=0.15]{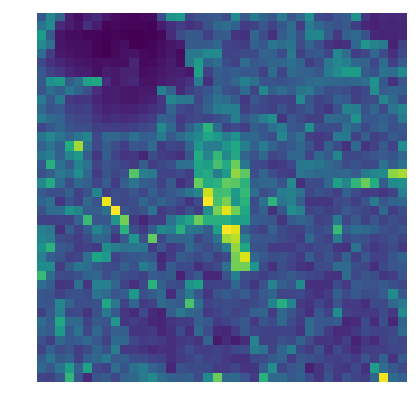}
\hspace{-0.1in}
\includegraphics[scale=0.15]{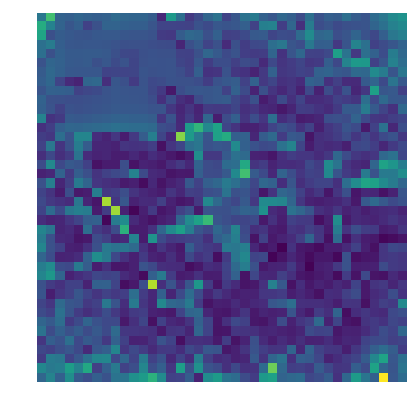}
\hspace{-0.1in}
\includegraphics[scale=0.15]{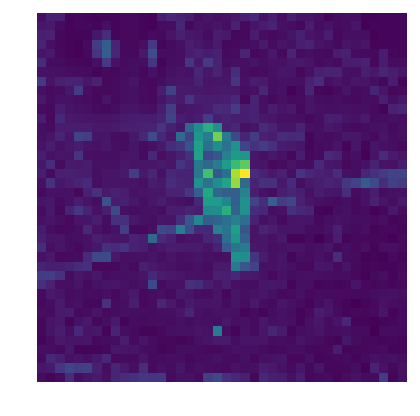}
\hspace{0.1in}
\includegraphics[scale=0.15]{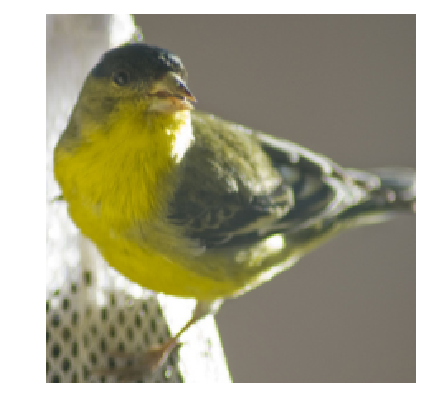}
\hspace{-0.1in}
\includegraphics[scale=0.15]{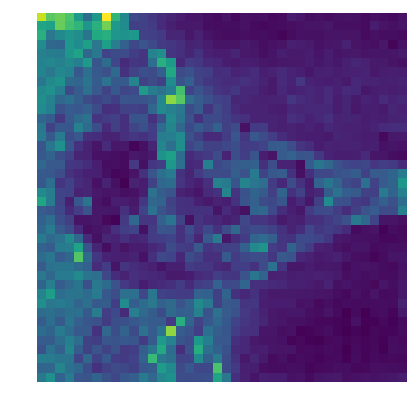}
\hspace{-0.1in}
\includegraphics[scale=0.15]{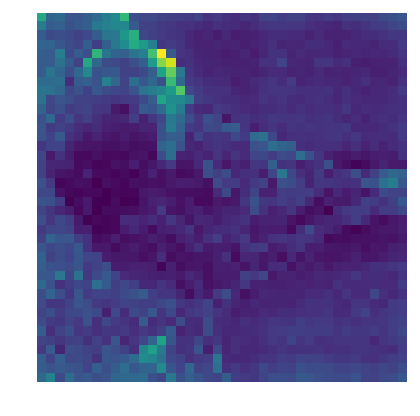}
\hspace{-0.1in}
\includegraphics[scale=0.15]{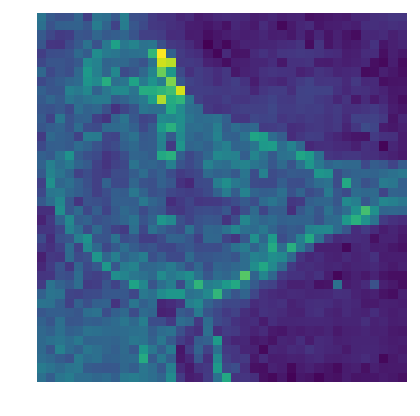}

\includegraphics[scale=0.15]{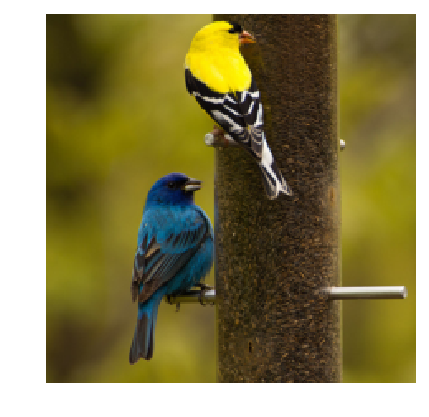}
\hspace{-0.1in}
\includegraphics[scale=0.15]{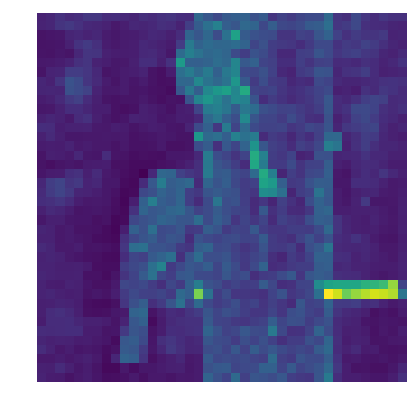}
\hspace{-0.1in}
\includegraphics[scale=0.15]{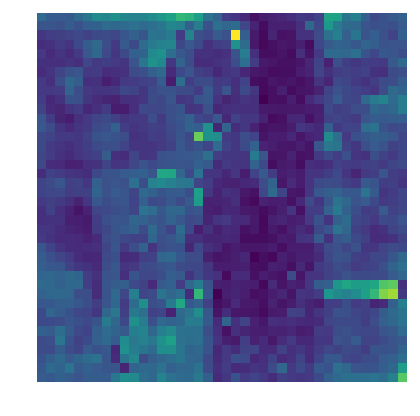}
\hspace{-0.1in}
\includegraphics[scale=0.15]{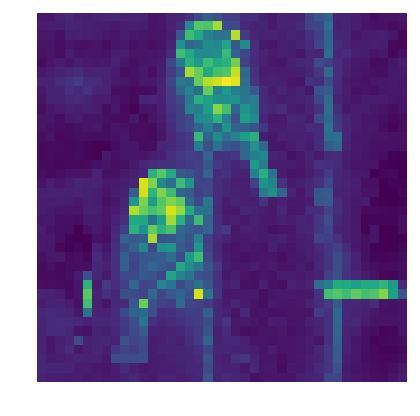}
\hspace{0.1in}
\includegraphics[scale=0.15]{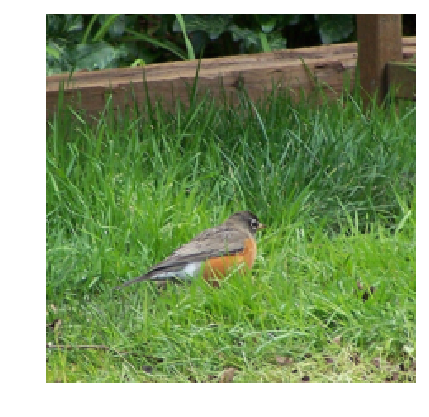}
\hspace{-0.1in}
\includegraphics[scale=0.15]{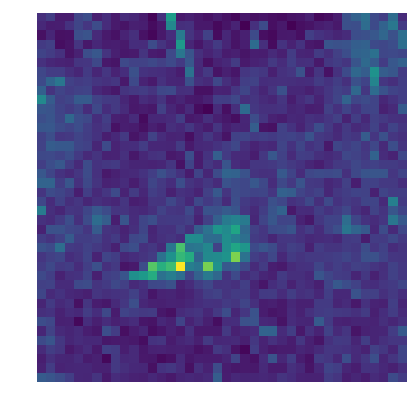}
\hspace{-0.1in}
\includegraphics[scale=0.15]{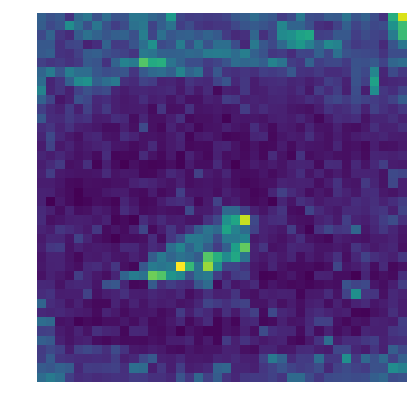}
\hspace{-0.1in}
\includegraphics[scale=0.15]{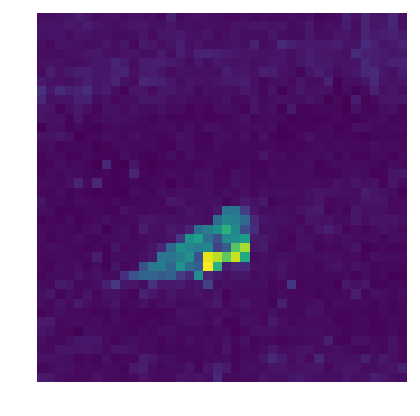}

\includegraphics[scale=0.15]{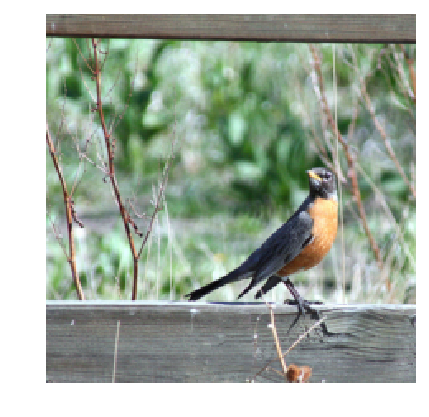}
\hspace{-0.1in}
\includegraphics[scale=0.15]{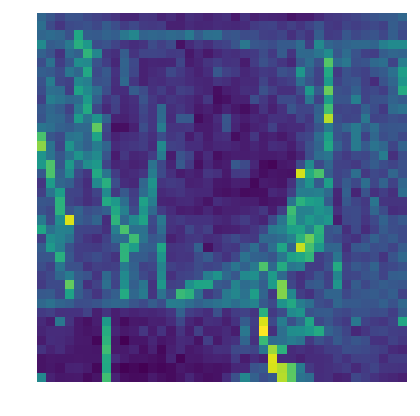}
\hspace{-0.1in}
\includegraphics[scale=0.15]{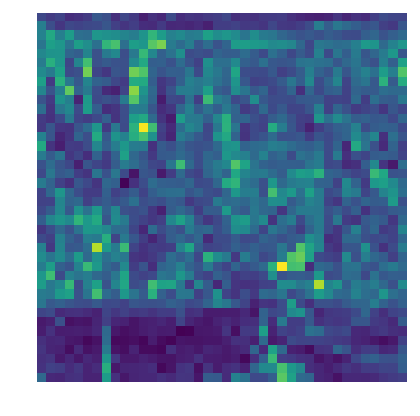}
\hspace{-0.1in}
\includegraphics[scale=0.15]{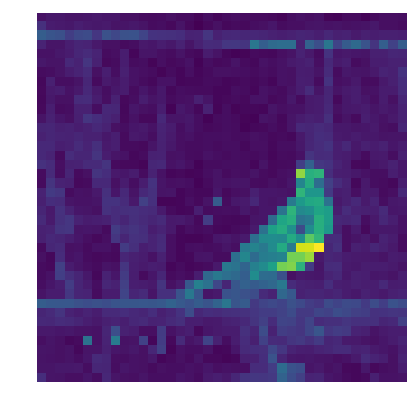}
\hspace{0.1in}
\includegraphics[scale=0.15]{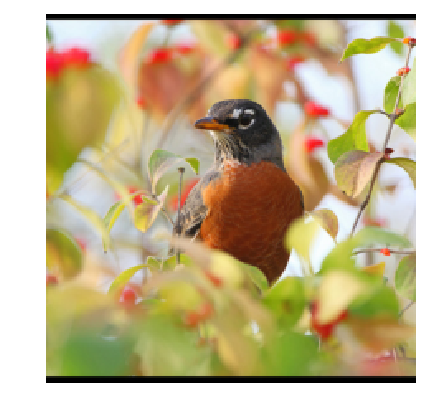}
\hspace{-0.1in}
\includegraphics[scale=0.15]{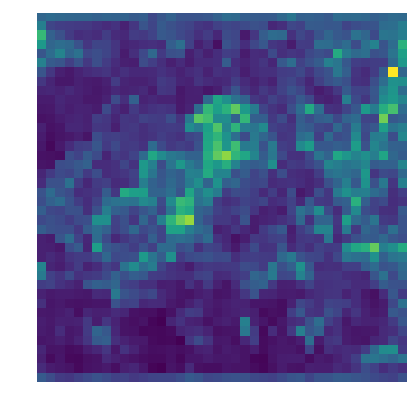}
\hspace{-0.1in}
\includegraphics[scale=0.15]{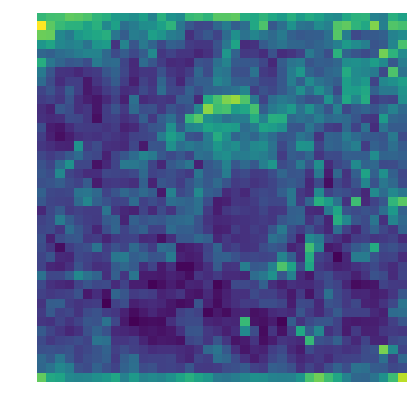}
\hspace{-0.1in}
\includegraphics[scale=0.15]{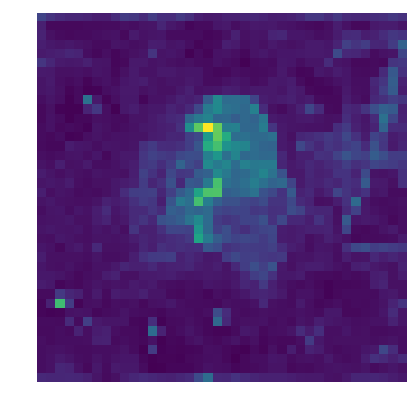}

\includegraphics[scale=0.15]{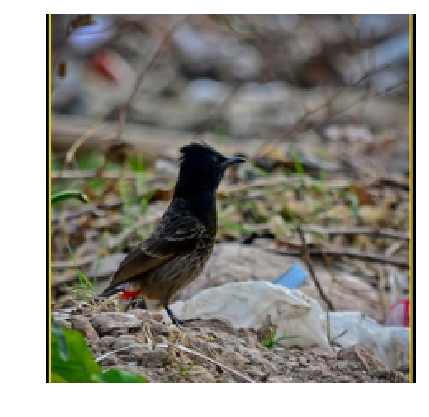}
\hspace{-0.1in}
\includegraphics[scale=0.15]{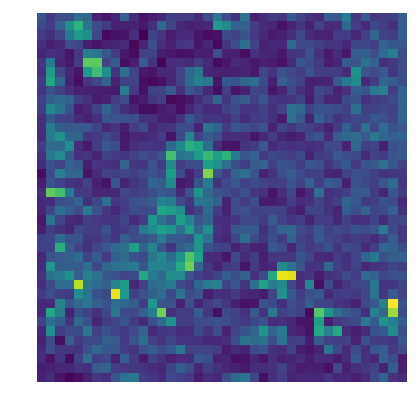}
\hspace{-0.1in}
\includegraphics[scale=0.15]{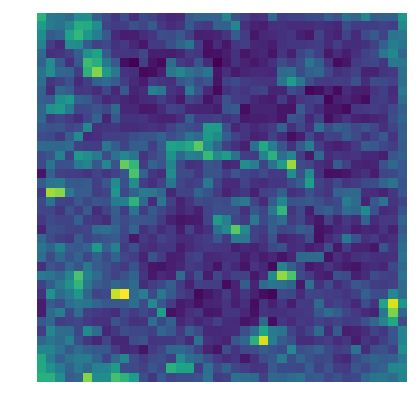}
\hspace{-0.1in}
\includegraphics[scale=0.15]{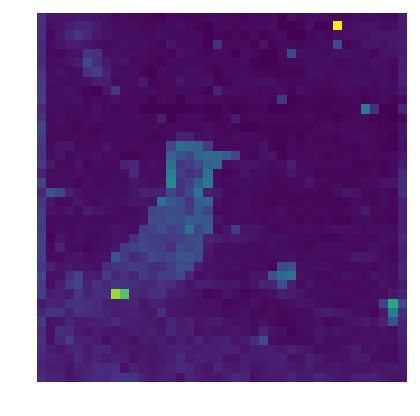}
\hspace{0.1in}
\includegraphics[scale=0.15]{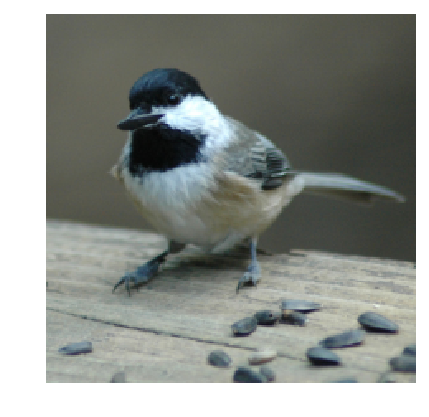}
\hspace{-0.1in}
\includegraphics[scale=0.15]{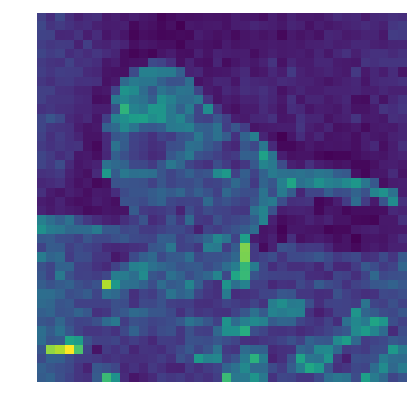}
\hspace{-0.1in}
\includegraphics[scale=0.15]{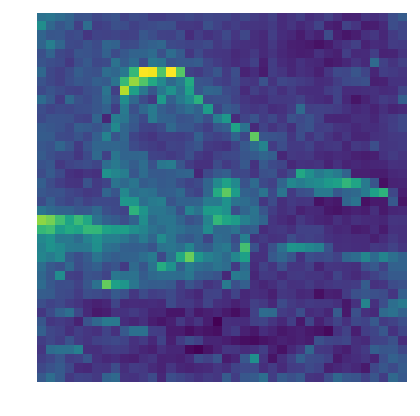}
\hspace{-0.1in}
\includegraphics[scale=0.15]{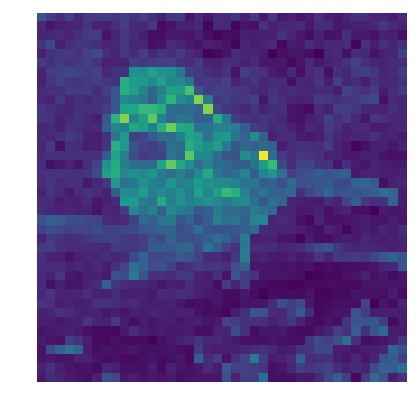}
\centerline{(a) \hspace{0.38in} (b) \hspace{0.38in} (c) \hspace{0.38in} (d) \hspace{0.55in} (a) \hspace{0.38in} (b) \hspace{0.38in} (c) \hspace{0.38in} (d)}
\caption{Visualizations of (a) original images, (b) the attention features extracted from the last self-attention layer of the ViT-Base model pretrained by MAE, (c) MAE with CLIP reconstruction target, and (d) our MILAN method. MILAN can better extract the important visual contents inside the images compared to both MAE and MAE with CLIP target.}
\label{fig:more_vis}

\hspace{0.1in}

\centering
\includegraphics[scale=0.15]{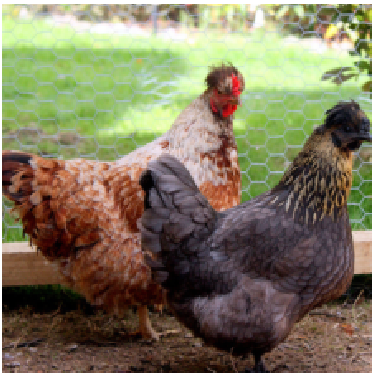}
\includegraphics[scale=0.15]{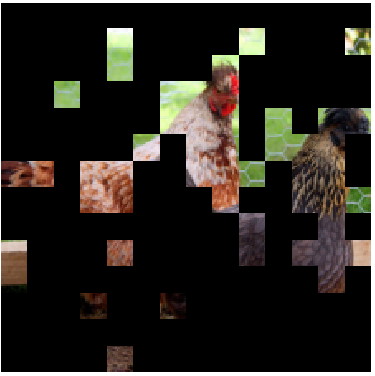}
\includegraphics[scale=0.134]{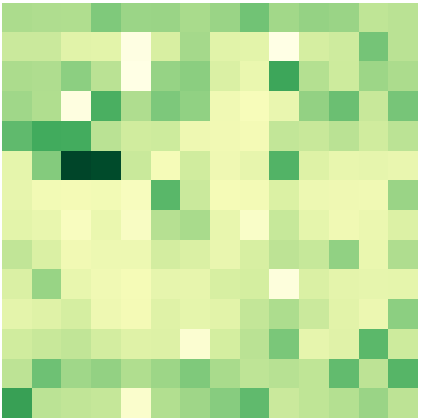}
\hspace{0.05in}
\includegraphics[scale=0.15]{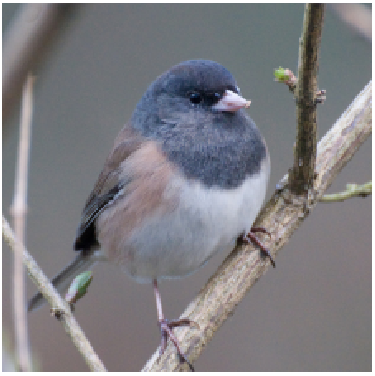}
\includegraphics[scale=0.15]{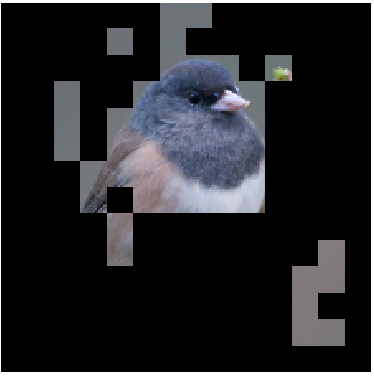}
\includegraphics[scale=0.134]{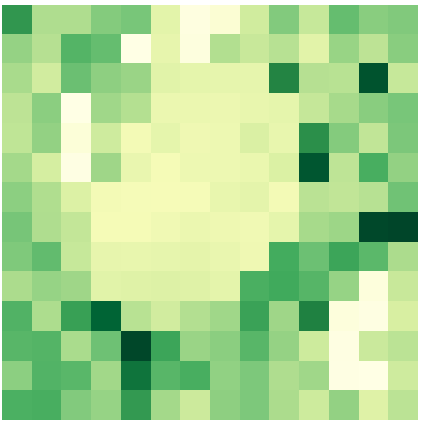}
\hspace{0.05in}
\includegraphics[scale=0.15]{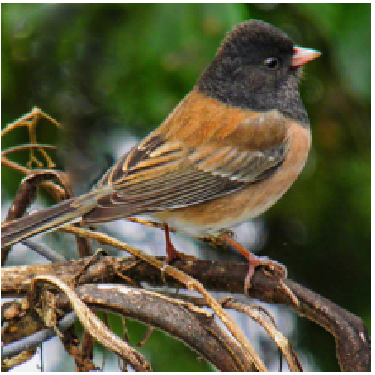}
\includegraphics[scale=0.15]{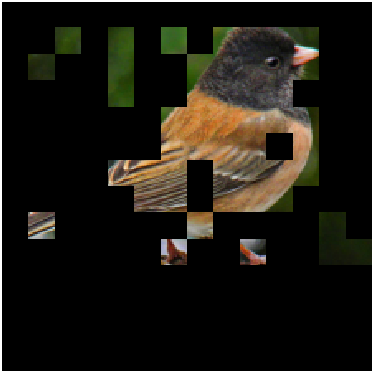}
\includegraphics[scale=0.134]{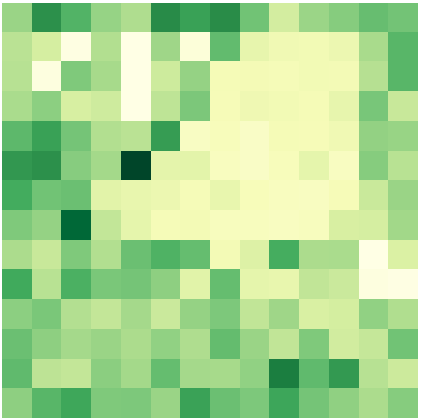}



\caption{Visualization of the original images (left), masked images by the semantic aware sampling strategy with 75\% masking ratio (middle), and the reconstruction loss patch-by-patch (right). For the plots of reconstruction loss, darker green colors indicate higher loss values. As shown, both unmasked patches and masked foreground patches have lower losses.}
\label{fig:ablation_mask}
\end{figure*}

\paragraph{Impact of different CLIP image encoders.}
In Table~\ref{tab:image_encoder}, we compare the results of using ViT-Base version and ViT-Large version of the CLIP image encoder to pretrain our ViT-Large model in the MILAN framework. The results are obtained by 400 epochs of pretraining on ImageNet-1K dataset, followed by finetuning, linear probing or transfer learning, using the same procedures as described in Appendix A.1.
Using image features produced from the ViT-Large CLIP image encoder as the targets consistently improves the performance on all tasks. For example, it achieves 1.1\% higher accuracy than using the ViT-Base CLIP image encoder on ImageNet finetuning.
Our improvements on ViT-Large are consistent with those on ViT-Base. For ViT-Large, MILAN achieved 87.8\% top-1 accuracy (2.4\% higher than our ViT-Base results) at 224x224 resolution and 88.3\% top-1 accuracy at 384x384 resolution on ImageNet. MILAN outperforms previous state-of-the-arts data2vec \cite{baevski2022data2vec} and PeCo \cite{dong2021peco} by 1.2\% and 1.3\%, respectively. The results clearly show that MILAN scales well with model sizes.

\subsection{Visualizations}
In Figure~\ref{fig:more_vis}, we provide visualizations of the learned representations from MAE, MAE+CLIP, and our MILAN method. MILAN can better extract the important visual contents inside the images compared to both MAE and MAE+CLIP, indicating that the proposed prompting decoder and semantic aware sampling contribute significantly to learning higher quality visual representations on top of the using the CLIP image features as reconstruction targets.
Moreover, in Figure~\ref{fig:ablation_mask}, we show that the proposed semantic aware sampling indeed favours more important image regions. The 25\% unmasked patches cover the contents that are more related to the objects in the images.
The semantic aware sampling facilitates the model to learn better on more important foreground regions, leading to accuracy improvements on finetuning, linear probing and semantic segmentation tasks as shown in our ablation studies.

\subsection{Limitation}
Similar to \cite{bao2021beit,chen2022cae,li2022mcbeit,wei2022mvp} which rely on external dataset{s} to train {their} image tokenizer{s}, the reconstruction target in MILAN is obtained from the CLIP model{ which also} requires an extra image-text dataset.
Training the CLIP model, if it is not amortized for many downstream tasks, is considered an extra training step.
However, in practice, we use publicly available pretrained CLIP models, so our method does not require a bespoke CLIP training step.
Moreover, we only perform inference on the CLIP image encoders to produce the reconstruction targets, which are only used in the pretraining phase. 
CLIP is not used in finetuning or linear probing stages, regardless of the classification, detection or segmentation tasks.
Although the feedfoward pass of the CLIP model incurs extra computation in the pretraining phase, we find that MILAN requires much less pretraining epochs compared to previous methods such as MAE \cite{he2021mae} and MaskFeat \cite{wei2021maskfeat}. For example, the actual wall-clock pretraining time of MILAN for ViT-Base is only half of the pretraining time of MAE.
Moreover, the language-image models like CLIP are becoming important and popular pretrained models to be applied to downstream tasks. Our method can be regarded as a useful intermediate step, given that our obtained models notably outperform these strong multi-modal models on various vision tasks.



\subsection{Societal impacts}
The proposed MILAN method produces transferable representations based on the learned statistics of the training dataset. Therefore, the trained model may also reflect the biases in the training data. Moreover, since MILAN uses the image features generated from the CLIP model as the reconstruction targets and the CLIP model itself is trained on an uncurated image-text dataset containing English-only captions, performance on images collected from non-English speaking countries requires further research. In future works, one {may} apply the MILAN method by taking multi-lingual language assisted representations as the reconstruction target. Extension to large-scale transformer model pretraining on video datasets using the MILAN framework could also be a future direction.

\end{document}